\PassOptionsToPackage{hyphens}{url}
\documentclass[runningheads]{llncs}
\usepackage{amsmath,amssymb}
\usepackage{lmodern}
\usepackage{ifxetex,ifluatex}
\ifnum 0\ifxetex 1\fi\ifluatex 1\fi=0 
  \usepackage[T1]{fontenc}
  \usepackage[utf8]{inputenc}
  \usepackage{textcomp} 
\else 
  \usepackage{unicode-math}
  \defaultfontfeatures{Scale=MatchLowercase}
  \defaultfontfeatures[\rmfamily]{Ligatures=TeX,Scale=1}
\fi
\IfFileExists{upquote.sty}{\usepackage{upquote}}{}
\IfFileExists{microtype.sty}{
  \usepackage[]{microtype}
  \UseMicrotypeSet[protrusion]{basicmath} 
}{}
\makeatletter
\@ifundefined{KOMAClassName}{
  \IfFileExists{parskip.sty}{%
    \usepackage{parskip}
  }{
    \setlength{\parindent}{0pt}
    \setlength{\parskip}{6pt plus 2pt minus 1pt}}
}{
  \KOMAoptions{parskip=half}}
\makeatother
\usepackage{xcolor}
\IfFileExists{xurl.sty}{\usepackage{xurl}}{} 

\usepackage{graphicx}
\makeatletter
\def\maxwidth{\ifdim\Gin@nat@width>\linewidth\linewidth\else\Gin@nat@width\fi}
\def\maxheight{\ifdim\Gin@nat@height>\textheight\textheight\else\Gin@nat@height\fi}
\makeatother
\setkeys{Gin}{width=\maxwidth,height=\maxheight,keepaspectratio}
\makeatletter
\def\fps@figure{htbp}
\makeatother
\providecommand{\tightlist}{%
  \setlength{\itemsep}{0pt}\setlength{\parskip}{0pt}}
\usepackage[utf8]{inputenc} 
\usepackage[T1]{fontenc}    
\usepackage{url}            
\usepackage{booktabs}       
\usepackage{amsfonts}       
\usepackage{nicefrac}       
\usepackage{microtype}      
\usepackage{graphicx}
\usepackage{doi}
\usepackage{mathtools}
\usepackage{bm} 
\usepackage{siunitx} 
\usepackage{dsfont}
\usepackage{float}
\usepackage{bookmark}

\makeatletter 
\renewcommand\@biblabel[1]{#1.}      
\makeatother

\usepackage{todonotes}

\newcommand{\funspace}{\mathcal{F}}
\newcommand{\embedspace}{\mathcal{Y}}
\newcommand{\pspace}{\Theta}

\newcommand{\co}{c}
\newcommand{\an}{a}
\newcommand{\Min}{\mathcal{M}_{\co}}
\newcommand{\Man}{\mathcal{M}_{\an}}
\newcommand{\Pin}{P_{\co}}
\newcommand{\Pan}{P_{\an}}
\newcommand{\phin}{\phi_{\co}}
\newcommand{\phan}{\phi_{\an}}
\newcommand{\Thin}{\Theta_{\co}}
\newcommand{\Than}{\Theta_{\an}}
\ifluatex
  \usepackage{selnolig}  
\fi
\usepackage[numbers,comma,sort,compress]{natbib}
\author{}
\date{\vspace{-2.5em}}

\begin{document}

\title{A geometric perspective on functional outlier detection}

%

\author{Moritz Herrmann\inst{1}\orcidID{0000-0002-4893-5812} \and Fabian Scheipl\inst{1}\orcidID{0000-0001-8172-3603}}
 \authorrunning{M. Herrmann \and F. Scheipl}

\institute{Department of Statistics, Ludwig-Maximilians-University\\ Ludwigstr. 33, 80539 Munich, Germany\\
\email{moritz.herrmann@stat.uni-muenchen.de}\\
\url{https://www.fda.statistik.uni-muenchen.de/index.html}}

\maketitle

\begin{abstract}
We consider functional outlier detection from a geometric perspective, specifically: for functional data sets drawn from a functional manifold which is defined by the data's modes of variation in amplitude and phase.
Based on this manifold, we develop a conceptualization of functional outlier detection that is more widely applicable and realistic than previously proposed.
Our theoretical and experimental analyses demonstrate several important advantages of this perspective:
It considerably improves theoretical understanding and allows to describe and analyse complex functional outlier scenarios consistently and in full generality, by differentiating between structurally anomalous outlier data that are off-manifold and distributionally outlying data that are on-manifold but at its margins.
This improves practical feasibility of functional outlier detection: We show that simple manifold learning methods can be used to reliably infer and visualize the geometric structure of functional data sets.
We also show that standard outlier detection methods requiring tabular data inputs can be applied to functional data very successfully by simply using their vector-valued representations learned from manifold learning methods as input features.
Our experiments on synthetic and real data sets demonstrate that this approach leads to outlier detection performances at least on par with existing functional data-specific methods in a large variety of settings, without the highly specialized, complex methodology and narrow domain of application these methods often entail.
\end{abstract}

\hypertarget{sec:intro}{%
\section{Introduction}\label{sec:intro}}

\hypertarget{sec:intro:problem}{%
\subsection{Problem setting and proposal}\label{sec:intro:problem}}

Outlier detection for functional data is a challenging problem due to
the complex and information-rich units of observations, which can be
``outlying'' or unusual in many different ways. Functional outliers are
often categorized into magnitude and shape outliers
\citep[e.g.]{dai2020functional, arribas2015discussion}, whereas Hubert
et al. \citep{hubert2015multivariate} differentiate between isolated and
persistent outliers, the latter further subdivided into shift, amplitude
and shape outliers. However, neither of these taxonomies yield precise,
explicit, fully general definitions, which makes it difficult to
theoretically describe, analyze and compare functional outliers.
Magnitude outliers, for example, have been defined as functional
observations ``outlying in some part or across the whole design domain''
\citep[p.~1]{dai2020functional} or as ``curves lying outside the range
of the vast majority of the data'' \citep[p.~2]{arribas2015discussion},
whereas Hubert et al. \citep[p.~3]{hubert2015multivariate} define
isolated outliers as observations which ``exhibit outlying behavior
during a very short time interval'', in contrast to persistent outliers
which ``are outlying on a large part of the domain''.

To cut through the confusion, we propose a geometric perspective on
functional outlier detection based on the well-known ``manifold
hypothesis'' \citep{ma2011manifold, lee2007nonlinear}. This refers to
the assumption that ostensibly complex, high-dimensional data lies on a
much simpler, lower dimensional manifold embedded in the observation
space and that this manifold's structure can be learned and then
represented in a low-dimensional space, often simply called embedding
space. We argue that such a perspective both clarifies and generalizes
the concept of functional outliers, without the need for any strong
assumptions or prior knowledge about the underlying data generating
process or its outliers. In terms of theoretical development, the
approach allows us to consistently formalize and systematically analyze
functional outlier detection in full generality. We also demonstrate
that procedures based on this perspective simplify and improve
functional outlier detection in practice: it suggests a principled, yet
flexible approach for applying well-established, highly performant
standard outlier detection methods such as local outlier factors (LOF)
\citep{breunig2000lof} to functional data, based on embedding
coordinates obtained via manifold learning or dimension reduction
methods. Our experiments show that doing so performs at least on par
with existing functional-data-specific outlier detection methods,
without the methodological complexity and limited applicability that
methods specific to functional data often entail. Moreover, such lower
dimensional representations serve as an easily accessible visualization
and exploration tool that helps to uncover complex and subtle data
structures which cannot be sufficiently reflected by one-dimensional
outlier scores or labels, nor captured by many of the previously
proposed 2D diagnostic visualizations for functional outliers.

\hypertarget{sec:prelims:background}{%
\subsection{Background and related work}\label{sec:prelims:background}}

\begin{figure}
\centering
\includegraphics{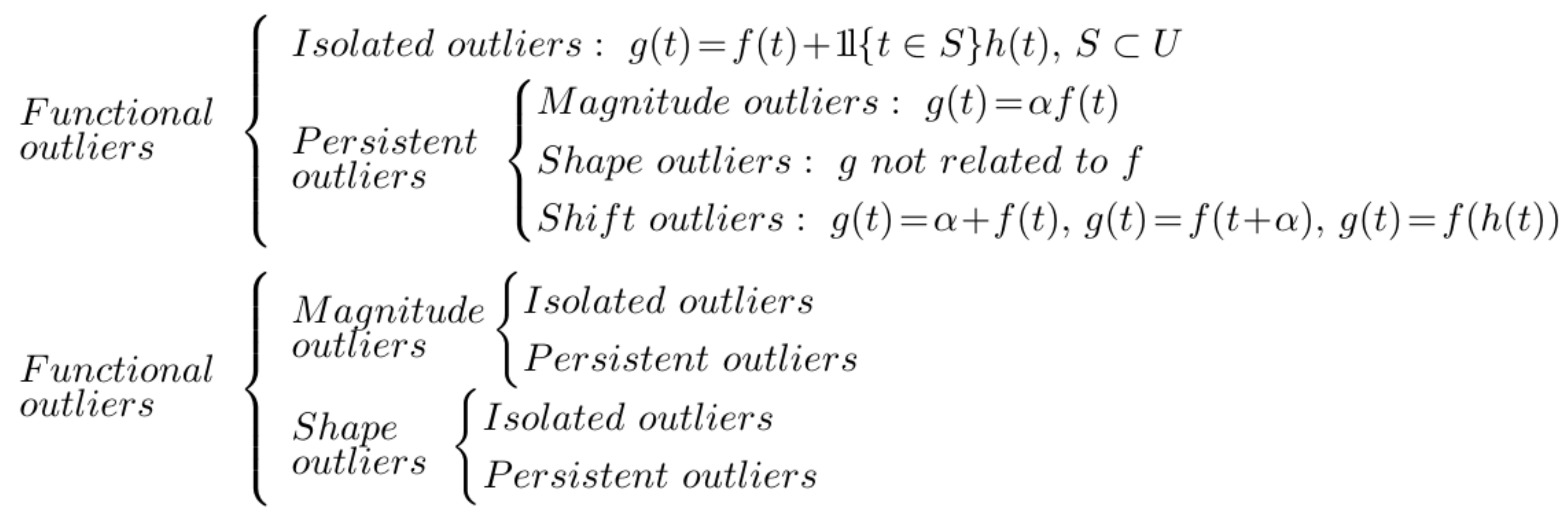}
\caption{Functional outlier taxonomies. Bottom: standard taxonomy. Top: taxonomy as introduced by Hubert et al.~Image taken from Arribas-Gil and Romo \citep{arribas2015discussion}. \label{fig:tax}}
\end{figure}

Functional data analysis (FDA) \citep[e.g.]{ramsay2005functional}
focuses on data where the units of observation are realizations of
stochastic processes over compact domains. In many cases, the intrinsic
dimensionality of functional data (FD) is much lower than the observed.
First, while FD are infinite dimensional in theory, they are
high-dimensional in practice -- functional observations are usually
recorded on fine and dense grids of argument values. Second, the
dominant drivers of differences between functional observations are
often comparatively low-dimensional so that just a few modes of
variation capture most of the structured variability in the data.\\
However, FD usually contain both amplitude and phase variation, i.e.,
``vertical'' shape or level variation as well as ``horizontal'' shape
variation. These different kinds of variability contribute to the
difficulty to precisely define and differentiate the various forms of
functional outliers and to develop methods that can ``catch them all'',
making outlier detection a highly investigated research topic in FDA.
For example, Arribas-Gil and Romo \citep{arribas2015discussion} argue
that the proposed outlier taxonomy of Hubert et al.
\citep{hubert2015multivariate} can be made more precise in terms of
expectation functions \(f(t)\) and \(g(t)\), with \(f(t)\) a ``common''
process, see Figure \ref{fig:tax}.

\begin{figure}
\includegraphics[width=1\linewidth]{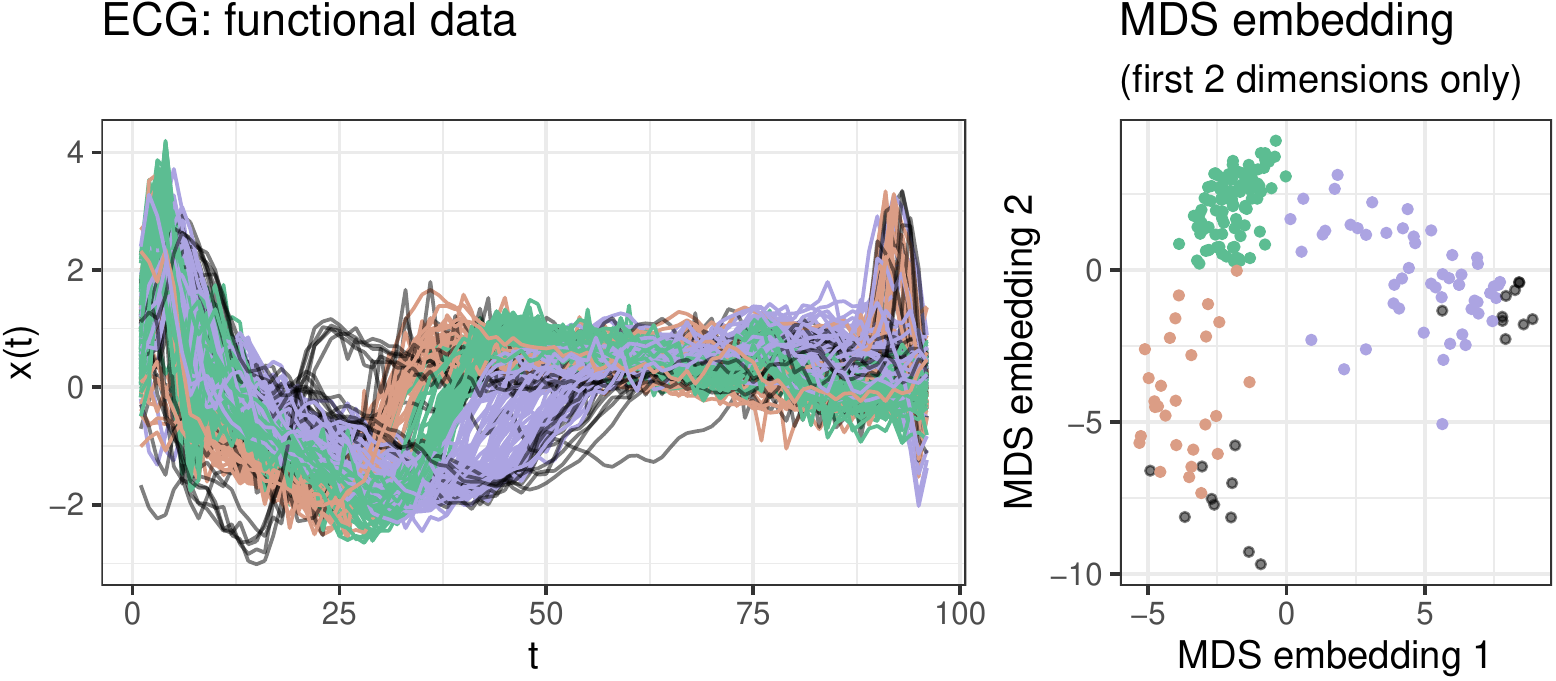} \caption{\label{fig:ecg}ECG curves and first two embedding dimensions (of five). Colors highlight subgroups apparent in the embeddings. Potential outliers with 5D-embedding LOF scores (minPts = $0.75n$) in top decile shown in black.}\label{fig:unnamed-chunk-10}
\end{figure}

Despite these attempts some fundamental issues remain unsolved. The
proposed taxonomies do not provide precise definitions and some of the
definitions are contradictory to some extent. Finally, many outlier
scenarios for realistic data generating processes are not covered by the
described taxonomies at all. As Arribas-Gil and Romo
\citep{arribas2015discussion} themselves point out, settings with
phase-varying data (i.e., ``horizontal'' variability through elastic
deformations of the functions' domains) are not sufficiently reflected,
as functions deviating in terms of phase may be considered as shape
outliers in cases where there are only few of such functions but not in
settings where all functions display such variation.\\
In addition, the taxonomy in Figure \ref{fig:tax} provides a reasonable
conceptual framework only if the non-outlying data from the ``common''
data generating process is characterized adequately just by its global
mean function. This cannot be assumed for many real data sets which
often contain highly variable sets of functions which display several
modes of phase, shape and/or amplitude variation simultaneously and/or
which come from multiple classes with class-specific means and higher
moments (see Figure \ref{fig:ecg}, e.g.).

Published research focuses mostly on the development of outlier
detection methods specifically for functional data, and a multitude of
methods based on a variety of different concepts such as functional data
depths \citep[e.g.]{hernandez2106kernel, harris2020elastic}, functional
PCA \citep{sawant2012fpca}, functional isolation forests
\citep{staerman2019functional}, robust functional archetypoids
\citep{vinue2020robust} or functional outlier metrics like directional
outlyingness \citep{rousseeuw2018measure, dai2019directional}, often
narrowly focused on detecting specific kinds of functional outliers,
have been put forth. Dai et al. \citep{dai2020functional} propose a
transformation-based approach to functional outlier detection and claim
that sequentially transforming shape outliers, which ``are much more
challenging to handle'', into magnitude outliers, makes them easier to
detect with established methods \citep[p.~2]{dai2020functional}. The
approach allows to define functional outliers more precisely in terms of
the transformations being used, like normalizing or centering functions
or taking their derivatives, but practitioners still need to be able to
come up with appropriate transformations for the data at hand first.\\
Recently, Xie et al. \citep{xie2017geometric} have introduced a
decomposition of functional observations into amplitude, phase and shift
components, based on which specific types of outliers can be identified
in a more general geometric framework without necessarily requiring
functional data to be of comparatively low rank. Similar in spirit to
our proposal, Hyndman and Shang \citep{hyndman2010rainbow} used kernel
density estimation and half-space depth contours of two-dimensional
robustified FPCA scores to construct functional boxplot equivalents and
detect outliers, and Ali et al. \citep{ali2019timecluster} use data
representations in two dimensions obtained from manifold methods for
outlier detection and clustering, but the focus of both is on
practicalities without considering the theoretical implications and
general applicability of embedding-based approaches nor do they consider
the necessity of higher dimensional representations.

The remainder of the paper is structured as follows: We provide the
theoretical formalization and discussion of the geometric approach in
section \ref{sec:theory}. Based on these theoretical considerations,
section \ref{sec:exps} presents extensive experiments. Section
\ref{sec:exps:qual-analysis} covers a detailed qualitative analysis of
real world ECG data, while section \ref{sec:exps:performance} provides
quantitative experiments and systematic comparisons to previously
proposed methods on complex synthetic outlier scenarios. We conclude
with a discussion in section \ref{sec:dis}.

\hypertarget{sec:theory}{%
\section{Functional outlier detection as a manifold learning
problem}\label{sec:theory}}

In this section, we first define two forms of functional outliers from a
geometrical view point: off- and on-manifold outliers. We then
illustrate how this perspective contains and extends existing outlier
taxonomies and how it can be used to formalize a large variety of
additional scenarios for functional data with outliers.

\hypertarget{the-two-notions-of-functional-outliers-off--and-on-manifold}{%
\subsection{The two notions of functional outliers: off- and
on-manifold}\label{the-two-notions-of-functional-outliers-off--and-on-manifold}}

\begin{figure}
\centering
\includegraphics{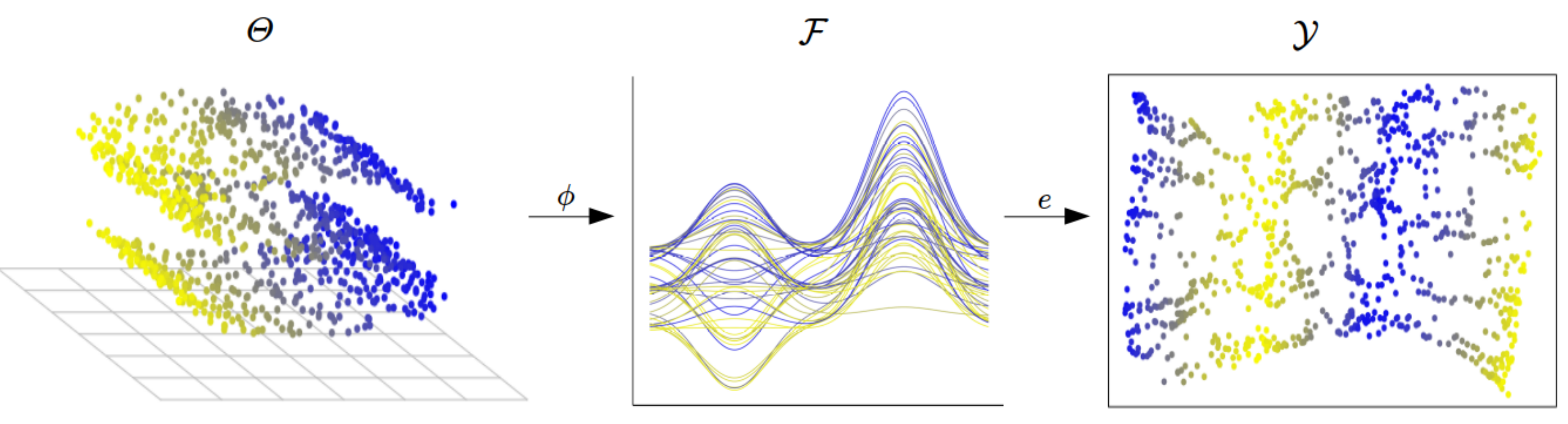}
\caption{Functional data from a manifold learning perspective.~Image taken from Herrmann \& Scheipl \citep{herrmann2020unsupervised}. \label{fig:framework}}
\end{figure}

Our approach to functional outlier detection rests on the \emph{manifold
assumption}, i.e., the assumption that observed high-dimensional data
are intrinsically low-dimensional. Specifically, we put forth that
observed functional data \(x(t) \in \funspace\), where \(\funspace\) is
a function space, arise as the result of a mapping
\(\phi: \pspace \to \funspace\) from a (low-dimensional) parameter space
\(\pspace \subset \mathbb{R}^{d_2}\) to \(\funspace\), i.e.,
\(x(t) = \phi(\theta)\). Conceptually, a \(d_2\)-dimensional parameter
vector \(\theta \in \pspace\) represents a specific combination of
values for the modes of variation in the observed functional data, such
as level or phase shifts, amplitude variability, class labels and so on.
These parameter vectors are drawn from a probability distribution \(P\)
over \(\mathbb{R}^{d_2}\):
\(\theta_i \sim P \;\forall\; \theta_i \in \pspace,\) with
\(\Theta = \{\theta : f_P(\theta) > 0\}\) and \(f_P\) the density to
\(P\). Mapping this parameter space to the function space creates a
functional manifold \(\mathcal{M}_{\pspace, \phi}\) defined by \(\phi\)
and \(\pspace\):
\(\mathcal{M}_{\pspace, \phi} = \{x(t): x(t) = \phi(\theta) \in \funspace, \theta \in \pspace\} \subset \funspace\),
an example is depicted in Figure \ref{fig:framework}. For
\(\funspace = L^2\) with data from a single functional manifold that is
isomorphic to some Euclidean subspace, Chen and Müller
\citep{chen2012nonlinear} develop a notion of a manifold mean and modes
of variation. Similarly, Dimelgio et al. \citep{dimeglio2014robust}
develop a robust algorithm for template curve estimation for connected
smooth sub-manifolds of \(\mathbb{R}^d\).

Unlike these single manifold settings, our conceptualization of outlier
detection is based on two functional manifolds. That is, we assume a
data set \(X = \{x_1(t), \dots, x_n(t)\}\) with \(n\) functional
observations coming from two separate functional manifolds
\(\Min = \mathcal{M}_{\Thin, \phin}\) and
\(\Man = \mathcal{M}_{\Than, \phan}\), with
\(\mathcal{M}_j \subset \funspace\), \(j \in \{\co, \an\}\) and
\(X \subset \{\Min \cup \Man\}\), with \(\Min\) representing the
``common'' data generating process and \(\Man\) containing anomalous
data. Moreover, for the purpose of outlier detection and in contrast to
the settings with a single manifold described in the referenced
literature, we are less concerned with precisely approximating the
intrinsic geometry of each manifold. Instead, it is crucial to consider
the manifolds \(\Min\) and \(\Man\) as sub-manifolds of \(\funspace\),
since we require not just a notion of distance between objects on a
single manifold, but also a notion of distance between objects on
different manifolds using the metric in \(\funspace\). Note that
function spaces such as \(\mathcal{C}\) or \(L^2\) which are commonly
assumed in FDA \citep{cuevas2014partial} are naturally endowed with such
a metric structure. Both, \(\mathcal{C}(D)\) and all \(L^p(D)\) spaces
over compact domain \(D\) are Banach spaces for \(p \geq 1\) and thus
also metric spaces \citep{malkowsky2019advanced}.\\
Finally, we assume that we can learn from the data an embedding function
\(e: \funspace \to \embedspace\) which maps observed functions to a
\(d_1\)-dimensional vector representation
\(y \in \mathcal{Y} \subset \mathbb{R}^{d_1}\) with \(e(x(t)) = y\)
which preserves at least the topological structure of \(\funspace\) --
i.e., if \(\Min\) and \(\Man\) are unconnected components of
\(\funspace\) their images under \(e\) are also unconnected in
\(\embedspace\) -- and ideally yields a close approximation of the
ambient geometry of \(\funspace\).

\textbf{Definition:} \emph{Off- and on-manifold outliers in functional
data}\\
Without loss of generality, let
\(r = \frac{\vert \{x_i(t): x_i(t) \in \Man\} \vert}{\vert \{x_i(t): x_i(t) \in \Min\} \vert} \lll 1\)
be the outlier ratio, i.e.~most observations are assumed to stem from
\(\Min\). Furthermore, let \(\Thin\) and \(\Than\) follow the
distributions \(\Pin\) and \(\Pan\), respectively. Let
\(\Omega^*_{\alpha, P}\) be an \(\alpha\)-minimum volume set of \(P\)
for some \(\alpha \in (0, 1)\), where \(\Omega^*_{\alpha, P}\) is
defined as a set minimizing the quantile function
\(V(\alpha) = \inf_{C \in \mathcal{C}}\{\text{Leb}(C): P(C) \geq \alpha\}, 0 < \alpha < 1\)\}
for i.i.d. random variables in \(\mathbb{R}^{d}\) with distribution
\(P\), \(\mathcal{C}\) a class of measurable subsets in
\(\mathbb{R}^{d}\) and Lebesgue measure \(\text{Leb}\)
\citep{polonik1997minimum}, i.e., \(\Omega^*_{\alpha, P}\) is the
smallest region containing a probability mass of at least \(\alpha\).

A functional observation \(x_i(t) \in X\) is then

\begin{itemize}
\tightlist
\item
  an off-manifold outlier if \(x_i(t) \in \Man\) and
  \(x_i(t) \notin \Min\).
\item
  an on-manifold outlier if \(x_i(t) \in \Min\) and
  \(\theta_i \notin \Omega^*_{\alpha, \Pin}\).
\end{itemize}

To paraphrase, we assume that there is a single ``common'' process
generating the bulk of observations on \(\Min\), and an ``anomalous''
process defining structurally different observations on \(\Man\). We
follow the standard notion of outlier detection in this, which assumes
that there are two data generating processes
\citep{ojo2021outlier, dai2020functional, zimek2018there}. Note this
does not necessarily imply that off-manifold outliers are similar to
each other in any way: \(\Pan\) could be very widely dispersed and/or
\(\Man\) could consist of multiple unconnected components representing
different kinds of anomalous data. The essential assumption here is that
the process from which \emph{most} of the observations are generated
yields structurally relatively similar data. This is reflected by the
notion of the two manifolds \(\Min\) and \(\Man\) and the ratio \(r\).
We consider settings with \(r \in [0, 0.1]\) as suitable for outlier
detection. By definition, the number of on-manifold outliers, i.e.,
\emph{distributional} outliers on \(\Min\) as opposed to the
\emph{structural} outliers on \(\Man\), only depends on the
\(\alpha\)-level for \(\Omega^*_{\alpha, \Pin}\).\\
Note that outlyingness in functional data is often defined only in terms
of shape or magnitude, but -- as we will illustrate in the following --
the concept ought to be conceived much more generally. The most
important aspect from a practical perspective is that structural
differences are reliably reflected in low dimensional representations
that can be learned via manifold methods, as we will show in Section
\ref{sec:exps}. These methods yield embedding coordinates
\(y \in \mathcal{Y}\) that capture the structure of data and its
outliers.

\hypertarget{sec:prelim:methods}{%
\subsection{Methods}\label{sec:prelim:methods}}

\begin{figure}[h]
\includegraphics[width=1\linewidth]{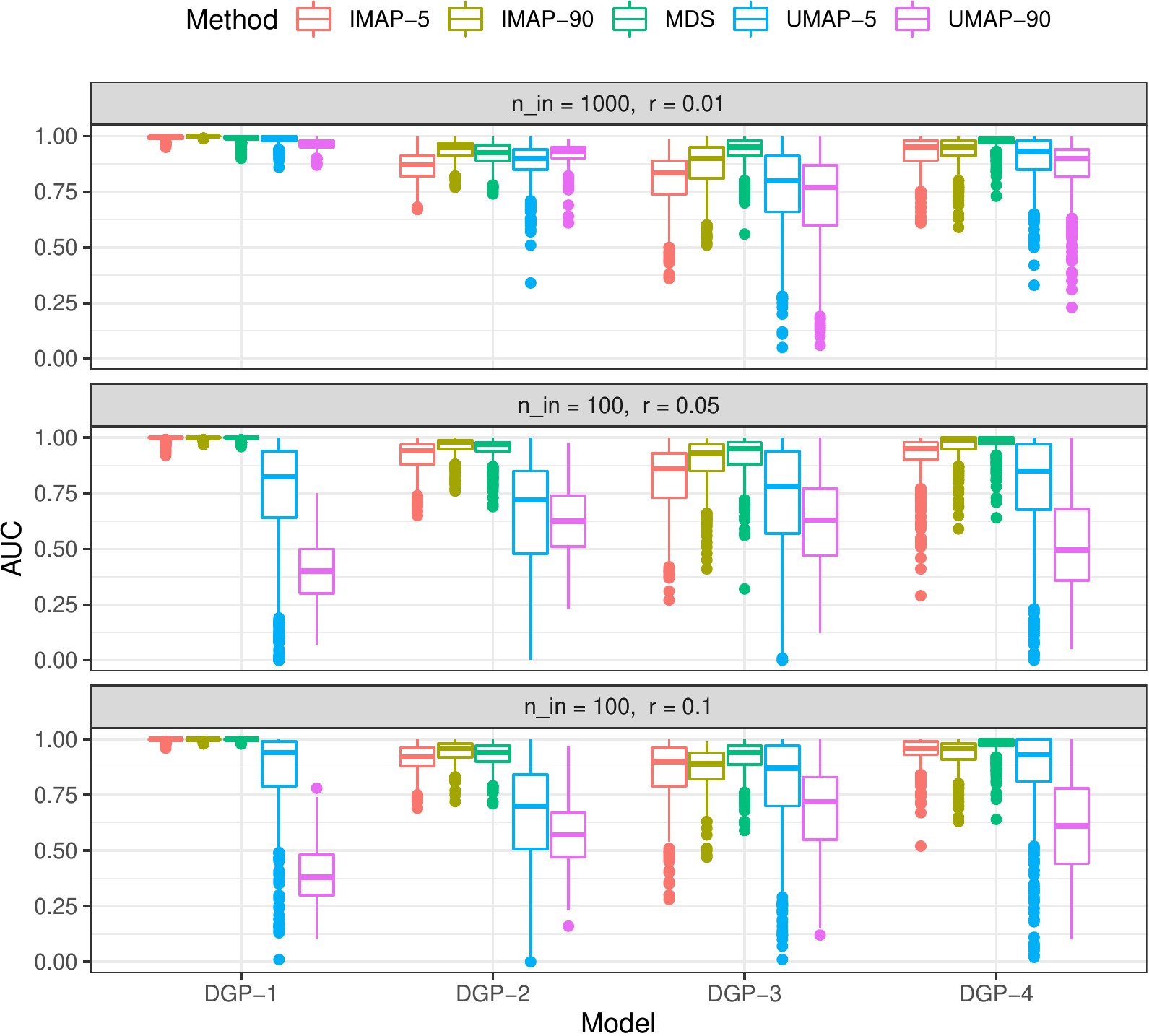} \caption{\label{fig:perf-uimap}Comparing UMAP and ISOMAP to MDS. UMAP and ISOMAP embeddings have been computed for two different locality parameter values: 5 and 90. Distribution of AUC over 500 replications of the four DGPs for different outlier ratios $r$. AUC computed on LOF scores based on 5D embeddings.}\label{fig:exp-uimap}
\end{figure}

To illustrate some of the implications of our general perspective on
functional outlier detection and showcase its practical utility, we
mostly use metric multi-dimensional scaling (MDS)
\citep{cox2008multidimensional} for dimension reduction and local
outlier factors (LOF) \citep{breunig2000lof} for outlier scoring in the
following. Note, however, that the proposed approach is not at all
limited to these specific methods and many other combinations of outlier
detection methods applied to lower dimensional embeddings from manifold
learning methods are possible. However, MDS and LOF have some important
favorable properties: First of all, both methods are well-understood,
widely used and tend to work reliably without extensive tuning since
they do not have many hyperparameters. Specifically, LOF only requires a
single parameter \(\verb|minPts|\) which specifies the number of nearest
neighbors used to define the local neighborhoods of the observations,
and MDS only requires specification of the embedding dimension.\\
More importantly, our geometric approach rests on the assumption that
functional outlier detection can be based on some notion of distance or
dissimilarity between functional observations, i.e., that abnormal or
outlying observations are separated from the bulk of the data in some
ambient (function) space. As MDS optimizes for an embedding which
preserves all pairwise distances as closely as possible (i.e., tries to
project the data \emph{isometrically}), it also retains a notion of
distance between unconnected manifolds in the ambient space. This
property of the embedding coordinates retaining the ambient space
geometry as much as possible is crucial for outlier detection. This also
suggests that manifold learning methods like ISOMAP
\citep{tenenbaum2000global}, t-SNE \citep{van2008visualizing} or UMAP
\citep{mcinnes2020umap}, which do not optimize for preservation of
\emph{ambient} space geometry via isometric embeddings by default, may
require much more careful tuning in order to be used in this way. Our
experiments support this theoretical consideration as can be see in
Figure \ref{fig:perf-uimap}. For LOF, this implies that larger values
for \(\verb|minPts|\) are to be preferred here, since such LOF scores
take into account more of the global ambient space geometry of the data
instead of only the local neighborhood structure. In Section
\ref{sec:exps}, we show that \(\verb|minPts| = 0.75n\), with \(n\) the
number of functional observations in a data set, seems to be a reliable
and useful default for the range of data sets we consider.\\
Two additional aspects need to be pointed out here. First, throughout
this paper we compute most distances using the \(L_2\) metric. This
yields MDS coordinates that are equivalent to standard functional PCA
scores (up to rotation). The proposed approach, however, is not
restricted to \(L_2\) distances. Combining MDS with distances other than
\(L_2\) yields embedding solutions that are no longer equivalent to PCA
scores, and suitable alternative distance measures may yield better
results in particular settings. We illustrate this aspect using the
\(L_{10}\) metric and two phase specific distance measures in section
\ref{sec:exps:general-dists}, which we apply to simulated data with
isolated outliers and a real data set of outlines of neolithic
arrowheads, respectively. Similarly, using alternative manifold learning
methods could be beneficial in specific settings, as long as they are
able to represent not just local neighborhood structure or on-manifold
geometry but also the global ambient space geometry.\\
Second, even though LOF could also be applied directly to the
dissimilarity matrix of a functional data set without an intermediate
embedding step, most anomaly scoring methods cannot be applied directly
to such distance matrices and require tabular data inputs. By using
embeddings that accurately reflect the (outlier) structure of a
functional data set, any anomaly scoring method requiring tabular data
inputs can be applied to functional data as well. In this work, we apply
LOF on MDS coordinates to evaluate whether functional data embeddings
can faithfully retain the outlier structure. Furthermore, embedding the
data before running outlier detection methods often provides large
additional value in terms of visualization and exploration, as the ECG
data analysis in Section \ref{sec:exps:qual-analysis} shows.

\hypertarget{examples-of-functional-outlier-scenarios}{%
\subsection{Examples of functional outlier
scenarios}\label{examples-of-functional-outlier-scenarios}}

We can now give precise formalizations of different functional outlier
scenarios and investigate corresponding low dimensional representations.
In this section we first show the geometrical approach is able to
describe existing taxonomies (see Figure \ref{fig:tax}) more
consistently and precisely. We then illustrate its ability to formalize
a much broader general class of outlier detection scenarios and discuss
the choice of distance metric and dimensionality of the embedding.

\hypertarget{outlier-scenarios-based-on-existing-taxonomies}{%
\subsubsection{\texorpdfstring{Outlier scenarios based on existing
taxonomies
\newline}{Outlier scenarios based on existing taxonomies }}\label{outlier-scenarios-based-on-existing-taxonomies}}

\begin{figure}
\includegraphics[width=1\linewidth]{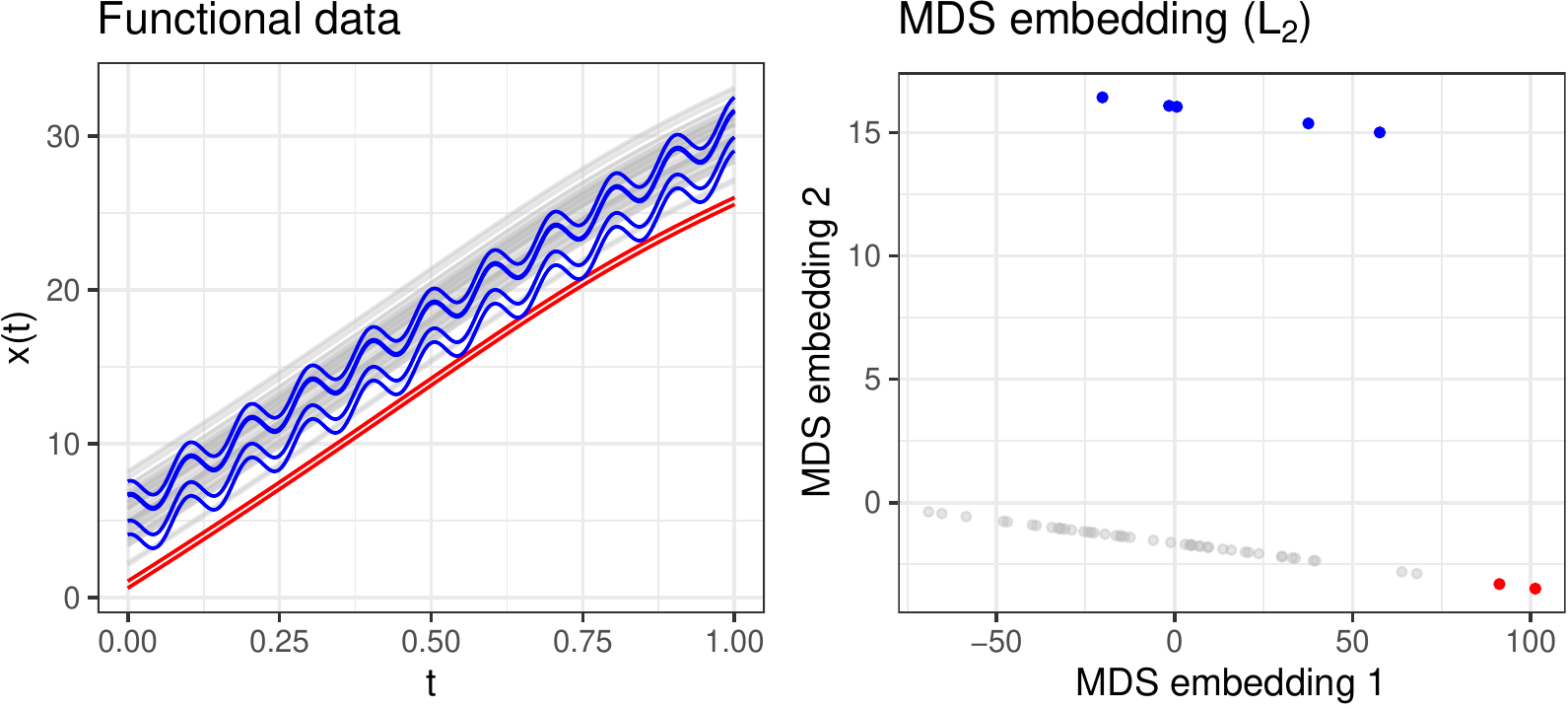} \caption{\label{fig:shape-example}Functional outlier scenario ($n = 54, r = .09$)  with shape variation inducing structural differences. Off-manifold outliers colored in blue, two on-manifold outliers colored in red.}\label{fig:unnamed-chunk-8}
\end{figure}

\textbf{Structure induced by shape:} In the taxonomy depicted in Figure
\ref{fig:tax}, top, the common data generating process is defined by the
expectation function \(f(t)\). This can be formalized in our geometrical
terms as follows: the set of functions defined by the ``common process''
\(f(t)\) defines a functional manifold (in terms of shape), i.e.~the
structural component is represented by the expectation function of the
common process. That means, we can define
\(\Min = \{x(t): x(t) = \theta f(t) = \phi(\theta, t), \theta \in \mathbb{R}\}\)
or
\(\Min = \{x(t): x(t) = f(t) + \theta = \phi(\theta, t), \theta \in \mathbb{R}\}\).
More generally, we can also model this jointly with
\(\Min = \{x(t): \theta_1 f(t) + \theta_2 = \phi(\mathbf{\theta}, t), \mathbf{\theta} = (\theta_1, \theta_2)' \in \mathbb{R}^2\}\).
In each case, magnitude and (vertical) shift outliers as defined in the
taxonomy correspond to on-manifold outliers in the geometrical approach,
as such observations are elements of \(\Min\). Isolated and shape
outliers, on the other hand, are by definition off-manifold outliers, as
long as ``\(g\) is not related to \(f\)'' is specified as
\(g \neq \theta f \,\forall\, \theta \in \mathbb{R}\). For example, if
we define \(\Man = \{x(t): x(t) = \theta g(t)\}\), it follows that
\(\Min \cap \Man = \emptyset\). The same applies to isolated outliers,
because \(g(t) = f(t) + I_U(t)h(t) \neq \theta_1 f(t) + \theta_2\).\\
Figure \ref{fig:shape-example} shows an example of such an outlier
scenario taken from \citep{hernandez2106kernel}. Following their
notation, the two manifolds can be defined as
\(\Min = \{x(t) \vert x(t) = b + 0.05t + \cos(20\pi t), b \in \mathbb R\}\)
and
\(\Man = \{x(t) \vert x(t) = a + 0.05t + \sin(\pi t^2), a \in \mathbb R\}\)
with \(t \in [0, 1]\) and \(a \sim N(\mu = 5, \sigma = 4)\),
\(b \sim N(\mu = 5, \sigma = 3)\). Note that the off-manifold outliers
lie within the mass of data in the visual representation of the curves,
whereas in the low dimensional embedding they are clearly separable.\\
However, we argue that the way \emph{shape} outliers are defined in
Figure \ref{fig:tax} is too restrictive, as many isolated outliers
clearly differ in shape from the main data, but are not captured by the
given definition if shape is considered in terms of ``\(g\) not related
to \(f\)''. In contrast, the geometrical perspective with its concepts
of off- and on-manifold outliers reflects that consistently. Another
issue with the considered taxonomy concerns horizontal shift outliers
\(f(t + \alpha)\) or \(f(h(t))\). Aribas-Gil and Romo
\citep{arribas2015discussion} specifically tackle that aspect in their
discussion. They distinguish between situations where ``all the curves
present horizontal variation'' (Case I), which is no outlier scenario
for them, and situations where only few phase varying observation are
present (Case II), which constitutes an outlier scenario. Again, the
geometric perspective allows to reflect that consistently. In appendix
\ref{sec:phase-var} we make these two notions explicit by defining
manifolds accordingly.

\hypertarget{general-functional-outlier-scenarios}{%
\subsubsection{\texorpdfstring{General functional outlier scenarios
\newline}{General functional outlier scenarios }}\label{general-functional-outlier-scenarios}}

\noindent As already noted, the concept of structural difference we
propose is much more general. It is straight forward to conceptualize
other outlier scenarios with induced structure beyond shape. Consider
the following theoretical example: Given a parameter manifold
\(\pspace \subset [0, \infty] \times [0, \infty] \times [0, \infty] \times [0, \infty]\)
and an induced functional manifold
\(\mathcal M = \{f(t); t \in [0, 1]: f(t) = \theta_1 + \theta_2 t^{\theta_3} + I(t \in [\theta_4 \pm .1])\}\).
Each dimension of the parameter space controls a different
characteristic of the functional manifold: \(\theta_1\) the level,
\(\theta_2\) the magnitude, \(\theta_3\) the shape, and \(\theta_4\) the
presence of an isolated peak around \(t = \theta_4\).\\
One can now define a ``common'' data generating process, i.e.~a manifold
\(\Min\), by holding some of the dimensions of \(\pspace\) fixed and
only varying the rest, either independently or not. On the other hand,
one can define an ``anomalous'' data generating process, i.e.~a
structurally different manifold \(\Man\), by letting those fixed in
\(\Min\) vary, or simply setting them to values unequal to those used
for \(\Min\), or by using different dependencies between parameters than
for \(\Min\). E.g., if \(\theta_1 = \theta_2\) for \(\Min\), let
\(\theta_1 = - \theta_2\) for \(\Man\). This implies one can define data
generating processes so that any functional characteristic (level,
magnitude, shape, ``peaks'' and their combinations) can be on-manifold
or off-manifold outliers, depending on how the ``common'' data manifold
\(\Min\) is defined.

\begin{figure}
\includegraphics[width=1\linewidth]{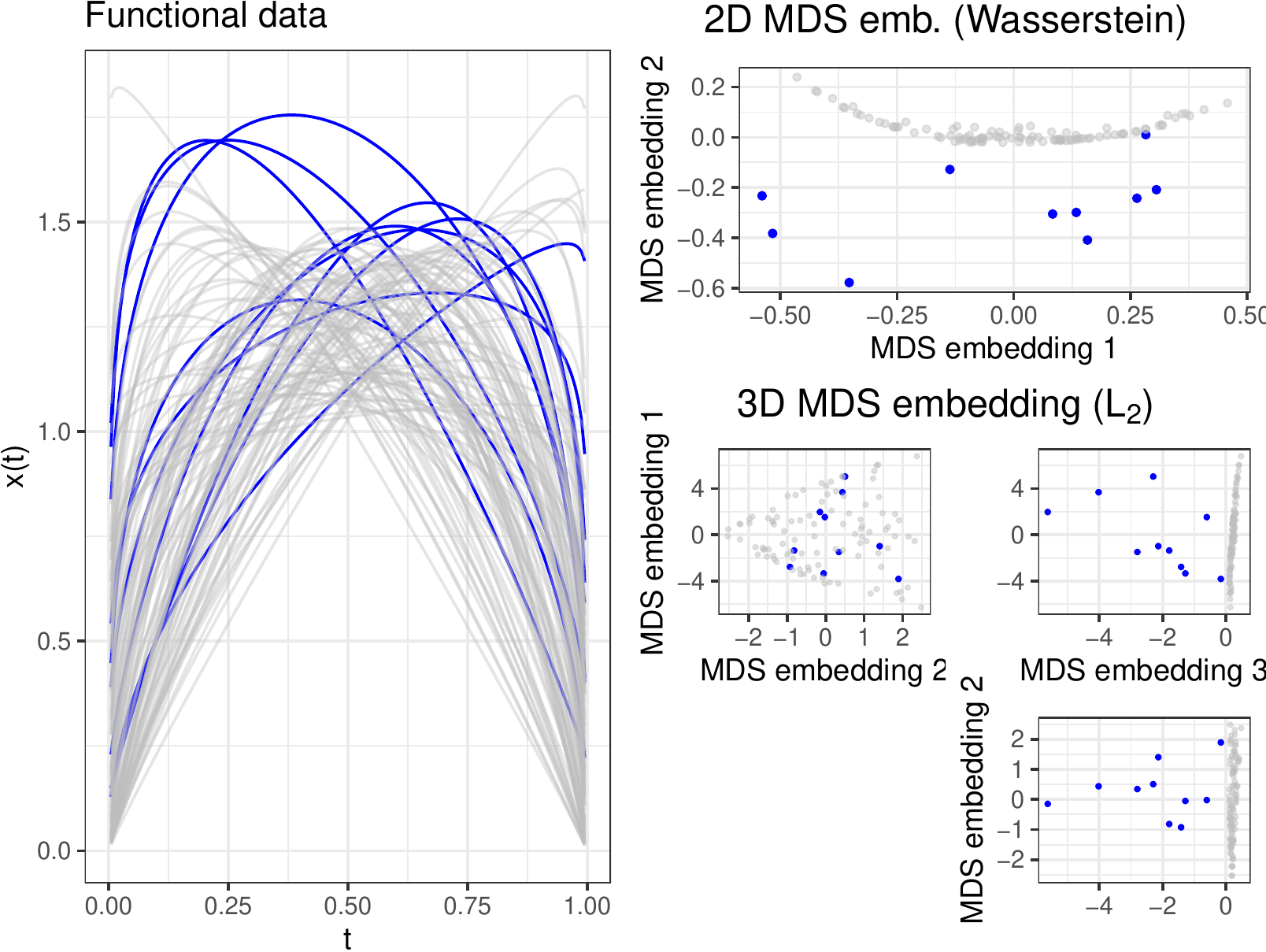} \caption{\label{fig:shift-example}Functional outlier scenario ($n = 100, r = .1$) with vertical shifts inducing structural differences. MDS embeddings based on unnormalized $L_1$-Wasserstein distances and $L_2$ (Euclidean) distances on the right.}\label{fig:theory-vert-shift}
\end{figure}

Figure \ref{fig:shift-example} shows a setting in which \(\Min\) is
defined purely in terms of complex shape variation while \(\Man\)
contains vertically shifted versions of elements in \(\Min\): Let
\(\Min\) be the functional manifold of Beta densities
\(f_{\text{B}}(t; \theta_1, \theta_2)\) with shape parameters
\(\theta_1, \theta_2 \in [1, 2]\), and let \(\Man\) be the functional
manifold of Beta densities with shape parameters
\(\theta_1, \theta_2 \in [1, 2]\) shifted vertically by some scalar
quantity \(\theta_3 \in [0, 0.5]\), that is
\(\Min = \{f(t); t \in [0, 1]: f(t) = f_{\text{B}}(t; \theta_1, \theta_2)\}\)
with \(\Thin = [1, 2]^2\) and
\(\Man = \{f(t); t \in [0, 1]: f(t) = f_{\text{B}}(t; \theta_1, \theta_2) + \theta_3\}\)
with \(\Than = \Thin \times [0, 0.5]\).

As can be seen in Figure \ref{fig:shift-example}, both manifolds contain
substantial shape variation that is identically structured, but those
from \(\Man\) are also shifted upwards by small amounts. Note that many
shifted observations lie within the main bulk of the data on large parts
of the domain. In the 2D embeddings based on unnormalized
\(L_1\)-Wasserstein distances \citep{gangbo2019unnormalized} (a.k.a.
``Earth Mover's Distance'', top right) and 3D embeddings based on
standard \(L_2\) distances (bottom right), we see that this structure is
captured with high accuracy, even though it is hardly visible in the
functional data, with most anomalous observations clearly separated from
the common manifold data, whose embeddings are concentrated on a narrow
sub-region of the embedding space. An observation on \(\Man\) which is
very close to \(\Min\), lying well within the main bulk of functional
observations, also appears very close to \(\Min\) in both embeddings.
This example shows that the two functional manifolds do not need to be
completely disjoint nor yield visually distinct observations for our
approach to yield useful results. It also shows that the choice of an
appropriate dissimilarity metric for the data can make a difference: a
2D embedding is sufficient for the more suitable Wasserstein distance
which is designed for (unnormalized) densities (top right panel), while
a 3D embedding is necessary for representing the relevant aspects of the
data geometry if the embedding is based on the standard \(L_2\) metric
(lower right panels). For a comparison with currently available outlier
visualization methods for this example, see Figure
\ref{fig:outliergram-ex5} in Appendix \ref{sec:compare-outliergram}.

In summary, we propose that the manifold perspective allows to define
and represent a very broad range of functional outlier scenarios and
data generating processes. We argue that these properties make the
geometrical approach very compelling for functional data, because it is
flexible, conceptualizes outliers on a much more general level (for
example, structural differences not in terms of shape) than before, and
allows to theoretically assess a given setting.\\
Beyond its theoretical utility of providing a general notion of
functional outliers, it has crucial practical implications: Outlier
characteristics of functional data, in particular structural
differences, can be represented and analysed using low dimensional
representations provided by manifold learning methods, regardless of
which functional properties define the ``common'' data manifold and
which properties are expressed in structurally different observations.
From a practical perspective, on-manifold outliers will appear
``connected'', whereas off-manifold outliers will appear ``separated''
in the embedding, and the clearer these structural differences are, the
clearer the separation in the embedding will be. Note that this implies
that shape outliers, which pose particular challenges to many previously
proposed methods, will often be particularly easily detectable.
Moreover, all methods for outlier detection that have been developed for
tabular data inputs can be (indirectly) applied to functional data as
well based on this framework, simply by using the embedding coordinates
as feature inputs: The embedding space \(\mathcal{Y}\) is typically a
low dimensional Euclidean space in which conventional outlier detection
works well and the essential geometrical structure encoded in the
pairwise functional distance matrix is conserved in these
lower-dimensional embeddings. In the next section, we illustrate this
practical utility in detail by extensive quantitative and qualitative
analyses.

\newpage

\hypertarget{sec:exps}{%
\section{Experiments}\label{sec:exps}}

To illustrate the practical relevance of the outlined geometrical
approach, we first qualitatively investigate real data sets. In the
second part of this section, we quantitatively investigate the anomaly
detection performance of several detection methods based on synthetic
data.

\hypertarget{sec:exps:qual-analysis}{%
\subsection{Qualitative analysis of real
data}\label{sec:exps:qual-analysis}}

\begin{figure}
\includegraphics[width=1\linewidth]{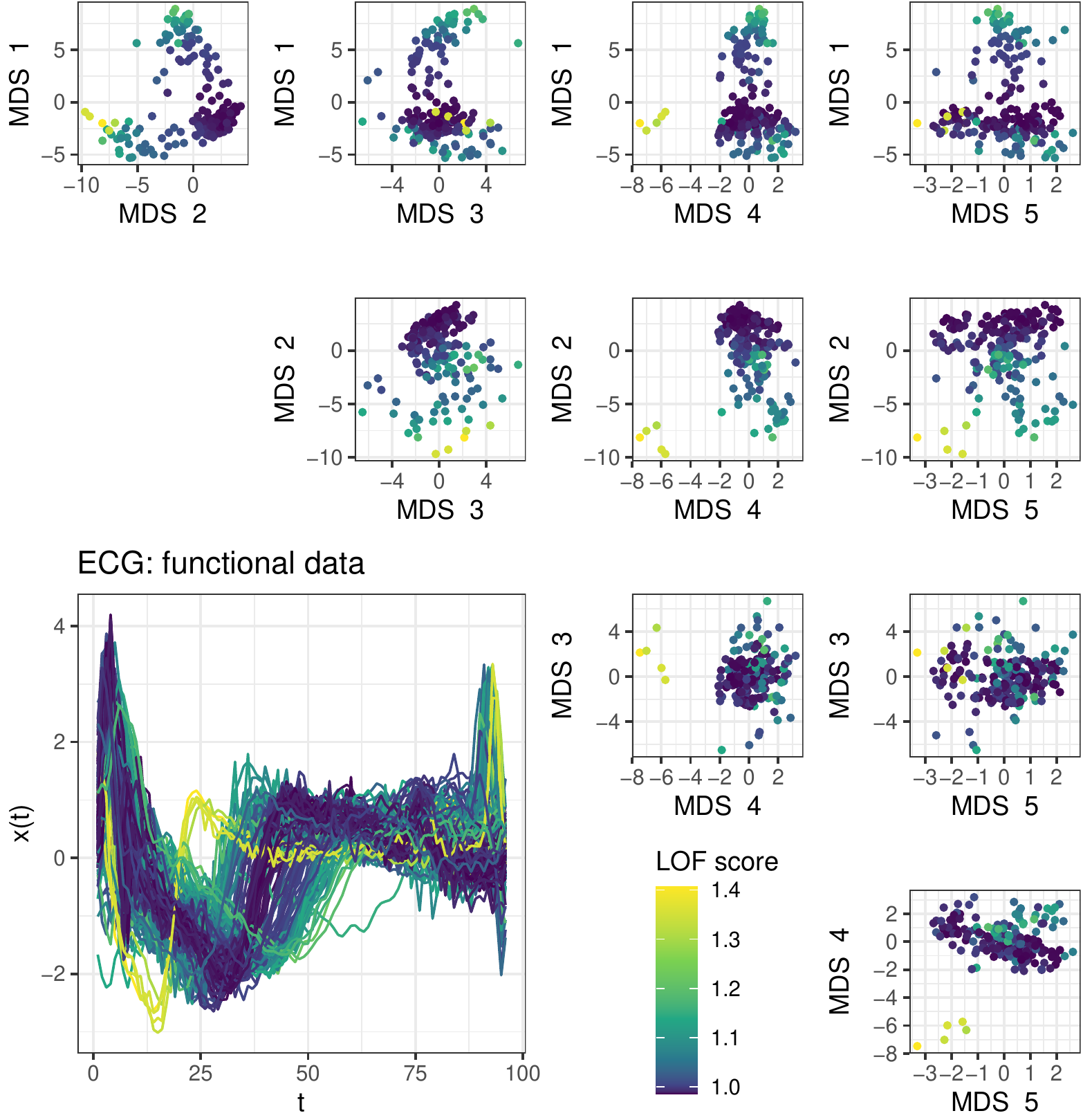} \caption{\label{fig:ecg-scatter}ECG data: Scatterplotmatrix of all 5 MDS embedding dimensions and curves, lighter colors for higher LOF score of 5D embeddings.}\label{fig:unnamed-chunk-11}
\end{figure}

We start with an in-depth analysis of the \textit{ECG200} data
\citep{bagnall2017great, olszewski2001generalized}, a functional data
set with complex structure: it seems to contain subgroups with phase and
amplitude variation and different mean functions. As a result, the data
set appears visually complex (Figure \ref{fig:ecg}, left). Without the
color coding it would be challenging to identify the three subgroups
(see lower left plot in Figure \ref{fig:ecg-scatter}). Moreover, there
are five left shifted observations (apparent at \(t \in [10, 25]\)) and
a single (partly) vertical shift outlier (apparent at
\(t \in [50, 75]\)) clearly detectable by the naked eye.

\begin{figure}
\includegraphics[width=1\linewidth]{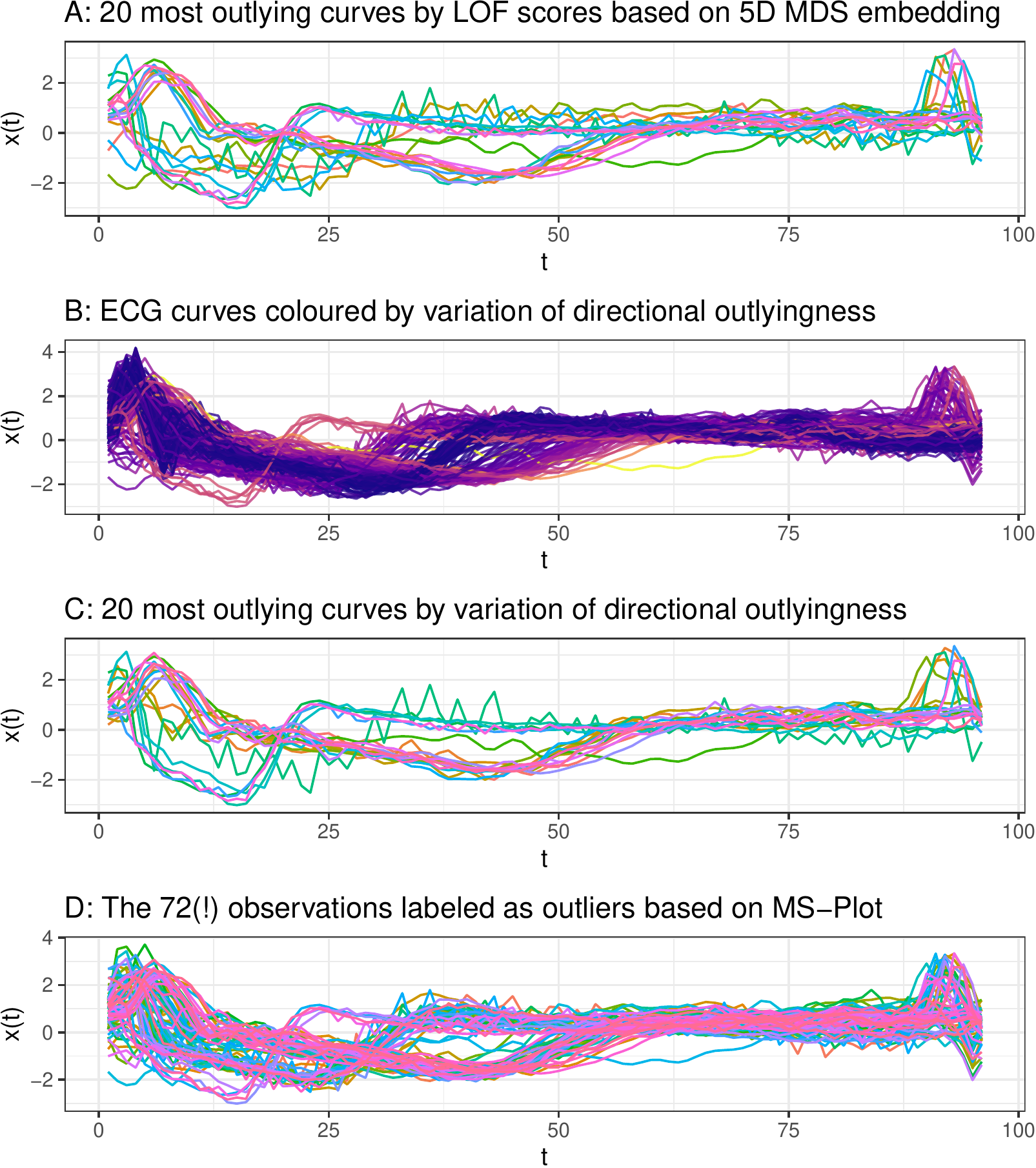} \caption{\label{fig:lof-vs-do}ECG data: LOF on MDS embeddings in contrast to directional outlyingness}\label{fig:unnamed-chunk-12}
\end{figure}

Much of the general structure (and the anomaly structure in particular)
becomes evident in a 5D MDS embedding. To begin with, in the first two
embedding dimensions, depicted on the right-hand side of Figure
\ref{fig:ecg}, three subgroups are easily recognizable. Color coding in
Figure \ref{fig:ecg} is based on this visualization. It makes apparent
that the substructures correspond to two smaller, horizontally shifted
subgroups of curves (red: left-shifted, purple: right-shifted), and a
central subgroup encompassing the majority of the observations (green).
In addition, we computed LOF scores on the 5D embedding coordinates. The
observations with LOF scores in the top decile are shown in black in
Figure \ref{fig:ecg}, two which the clearly outlying observations
belong.\\
More importantly, note that these observations are clearly separated
from the rest in a 5D embedding: the five clearly left shifted
observations in the fourth embedding dimension and the single vertically
shifted observation in the subspace spanned by the first and third
embedding dimension: Figure \ref{fig:ecg-scatter} shows a scatterplot
matrix of all five embedding dimensions with observations color-coded
according to the 5D-embedding LOF scores. The clear left shifted
outliers obtain the highest LOF scores due their isolation in the
subspaces including the fourth embedding dimension. Note, moreover, that
other observations with higher LOF scores appear in peripheral regions
of the different subspaces, but they are not as clearly separable as the
six observations described before. Regarding Figure \ref{fig:lof-vs-do}
A, which shows the 20 most outlying curves according to the LOF scores,
this can be explained by the fact that these other observations stem
from one of the two shifted subgroups and can thus be seen as
on-manifold outliers, whereas the six other, visually clearly outlying
observation are clear off-manifold outliers.

We contrast these findings with the results of directional outlyingness
\citep{dai2018multivariate, dai2019directional}, which performs very
well (see section \ref{sec:exps:performance}) on simple synthetic data
sets. Figure \ref{fig:lof-vs-do} shows the ECG curves color-coded by
variation of directional outlyingess (B), the 20 most outlying curves by
variation of directional outlyingness (C) and the observations labeled
as outliers by directional outlyingness respectively by the MS-plot (D).
First of all, it can be seen that many observations yield high variation
of directional outlyingness and observations in the right shifted
subgroup obtain most of the highest values. In fact, among the 20
observations with highest variation of directional outlyingness, only
one is from the left shifted group and 13 are from the right-shifted
group. Moreover, applying directional outlyingness to this data set
results in 72 observations being labeled as outliers, which is about 36
percent of all observations. We would argue that it is questionable
whether 36 percent of all observations should be labeled as outliers.\\
In this regard, the ECG data serves as an example which illustrates the
advantages of the geometric approach. First of all, it yields readily
available visualizations, which reflect much more of the inherent
structure of a data set than only anomaly structure. This is
specifically important for data with complex structure (i.e., subgroups
or multiple modes and large variability). Moreover, it allows to apply
well-established and powerful outlier scoring methods like LOF to
functional data. This exemplifies that the approach not only improves
theoretical understanding and consideration as outlined in the previous
section, it also has large practical utility in complex real data
settings in which previously proposed methods may not provide useful
answers.

\begin{figure}
\includegraphics[width=1\linewidth]{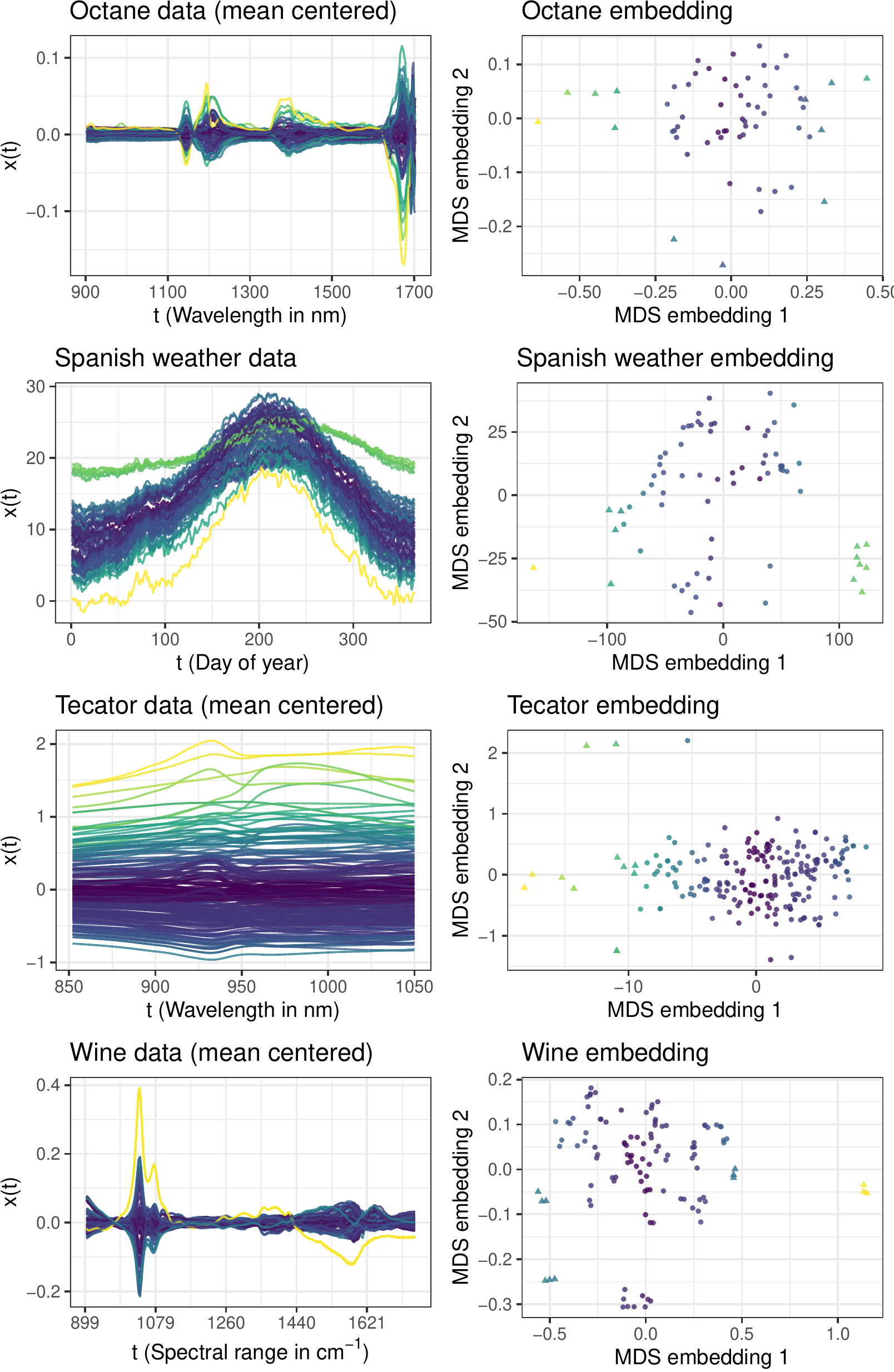} \caption{\label{fig:real-dat}Further examples of real functional data colored by LOF score. The 12 most outyling observations depicted as triangles in the embedding.}\label{fig:unnamed-chunk-14}
\end{figure}

In the ECG example, we have seen that a 5D embedding yielded reasonable
results and sufficiently reflected many aspects of the data. In
particular, the extremely left-shifted observations became clearly
separable in the 4th embedding dimension. In appendix \ref{sec:sms} we
analyse a synthetic data set in the same way as the ECG data, which
yields similar findings. Moreover, note that the Spearman rank
correlation between LOF scores computed on the 5D embedding and LOF
scores computed directly on the ECG data distances is 0.99. This shows
that outlier structure retained in the 5D embedding is highly consistent
with the outlier structure in the high dimensional observation space, an
important aspect with respect to anomaly scoring methods requiring (low
dimensional) tabular inputs.\\
Finally, note that even fewer than five embedding dimensions may suffice
to reflect much of the inherent structure. Consider the examples
depicted in Figure \ref{fig:real-dat}, which shows the functional
observations and the first two embedding dimensions of a corresponding
5D MDS embedding of another four real data sets. The Octane data consist
of spectra from 60 gasoline samples \citep{fds, kalivas1997two}, the
Spanish weather data of annual temperature curves of 73 weather stations
\citep{febrero2012fdausc}, the Tecator data of spectrometric curves of
meat samples \citep{febrero2012fdausc, ferraty2006nonparametric}, and
the Wine data of spectrometric curves of wine samples
\citep{holland1998use, bagnall2017great}. As before, the observations
are colored according to LOF scores based on the 5D embedding. In
addition, the 12 observations with highest LOF scores are depicted as
triangles. These data sets are much simpler than the ECG data and the
first two embedding dimensions already reflect the (outlier) structure
fairly accurately: observations with high LOF scores appear separated in
the first two embedding dimensions and more general substructures are
revealed as well. The substructure of the weather data is rather obvious
already regarding the functional observations, for example, the
observations with less variability in terms of temperature, all of which
obtain high LOF scores. The substructure of the wine data -- for example
the small cluster in the lower part of the embedding -- is much harder
to detect based on visualizations of the curves alone. Figure
\ref{fig:outliergram-real} in appendix \ref{sec:compare-outliergram}
shows results for the ``Outliergram'' by Aribas-Gil \& Romo
\citep{arribas2014shape} for shape outlier detection as well as the
magnitude-shape plot method of Dai \& Genton \citep{dai2018multivariate}
for these example data sets for comparison. We would argue that the
embedding based visualizations offers a much more informative
visualization of the structure of these data.\\
Appendix \ref{sec:exps:sensitivity} summarizes a more detailed analysis
of the sensitivity of the approach to the choice of the dimensionality
of the embedding. We conclude that sensitivity seems to be fairly low.
For all 5 real data sets we consider, the rank order of LOF scores is
very similar or even identical whether based on 2, 5 or even 20
dimensional embeddings (c.f. Table \ref{tab:cor}).\\
Following Mead \citep{mead1992review}, we quantify the goodness of fit
(GOF) for a \(d_1\)-dimensional MDS embedding as
\(GOF(d_1) = \frac{\sum_{i=1}^{d_1} \max(0, \lambda_i)}{\sum_{j = 1}^n \max(0, \lambda_j)},\)
where \(\lambda_k\) are the eigenvalues (sorted in decreasing order) of
the \(k\)th eigenvectors of the centered distance matrix. For all of the
considered real data sets, a 5D embedding achieved a goodness of fit
over 0.8, the four less complex examples even over 0.95 (see Figure
\ref{fig:expl-variation}). As a rule of thumb, the embedding dimension
does not seem crucial as long as the goodness of fit (GOF) of the
embedding is over 0.8 for \(L_2\) distances. This rule of thumb also
yields compelling quantitative performance results, as shown in section
\ref{sec:exps:performance}.

\newpage

\hypertarget{sec:exps:performance}{%
\subsection{Quantitative analysis of synthetic
data}\label{sec:exps:performance}}

In this section we investigate the outlier detection performance
quantitatively, based on synthetic data sets for which the true
(outlier) structure is known.

\hypertarget{sec:exps:performance:meths}{%
\subsubsection{Methods}\label{sec:exps:performance:meths}}

In addition to applying LOF to 5D embeddings and directly to the
functional data, we investigate the performance of two ``functional
data''-specific outlier detection methods: directional outlyingness (DO)
\citep{dai2018multivariate, dai2019directional} and total variational
depth (TV) \citep{huang2019decomposition}. We use implementations
provided by package \texttt{fdaoutlier} \citep{fdaoutlier} and use the
variation of directional outlyingness as returned by function
\(\verb|dir_out|\) as outlier scores for DO and the total variation
depths as returned by function \(\verb|total_variation_depth|\) for TV.

\hypertarget{sec:exps:performance:mods}{%
\subsubsection{Data Generating
Processes}\label{sec:exps:performance:mods}}

The methods are applied to data from four different data generating
processes (DGPs), the first two of which are based on the simulation
models introduced by Ojo et al. \citep{ojo2021outlier} and provided in
the corresponding R package \(\verb|fdaoutlier|\) \citep{fdaoutlier}. We
also provide the results of additional experiments based on the original
DGPs from package \texttt{fdaoutlier} in Appendix
\ref{sec:quant-fdaoutlier}. However, we consider most of these DGPs as
too simple for a realistic assessment, as most methods achieve almost
perfect performance on them and we use more complex DGPs here. In both
DGP 1 and 2, the inliers from \(\verb|simulation_model1|\) from package
\texttt{fdaoutlier} serve as \(\Min\), i.e.~the common data generating
process. This results in simple functional observations with a positive
linear trend. In addition, \(\verb|simulation_model1|\) generates simple
shift outliers. Additionally, our DGP 1 also includes shape outliers
stemming from \(\verb|simulation_model8|\) which serves as \(\Man\). In
contrast, DGP 2 contains shape outliers from all of the other DGPs in
\texttt{fdaoutlier}, which means \(\Man\) contains observations from
several different data generating processes.\\
For DGPs 3 and 4, we define \(\Min\) by generating a random, wiggly
template function over \([0, 1]\) for each data set, generated from a
B-spline basis with 15 or 25 basis functions, respectively, with i.i.d.
\(\mathcal N(0,1)\) spline coefficients. Functions in \(\Min\) are
generated as elastically deformed versions of this template, with random
warping functions drawn from the ECDFs of \(\text{Beta}(a, b)\)
distributions with \(a, b \sim U[4, 6]\) (DGP 3) or
\(a, b \sim U[3, 8]\) (DGP 4). Functions in \(\Man\) are also generated
as elastically deformed versions of this template, with Beta ECDF random
warping functions with \(a, b \sim U[3, 4]\) for DGP 3 and with 50:50
Beta mixture ECDF random warping functions with
\(a, b \sim 0.5 U[3, 8]:0.5 U[0.1, 3]\) (DGP 4). Finally, white noise
with \(\sigma = 0.1, 0.15\), respectively, for DGP 3 and 4 is added to
all resulting functions. Appendix \ref{app:dgp-ex} shows visualizations
of example data sets drawn from these DGPs.

\newpage

\hypertarget{sec:exps:performance:perfs}{%
\subsubsection{Performance
assessment}\label{sec:exps:performance:perfs}}

From these four DGPs we sampled data \(B = 500\) times with three
different outlier ratios \(r \in \{0.1, 0.05, 0.01\}\). Based on the
outlier scores, we computed the Area-under-the-ROC-Curve (AUC) as
performance measure and report the results over all 500 replications.
Note that, for \(r \in \{0.1, 0.05\}\), the number of sampled
observations is \(n = 100\), whereas for \(r = 0.01\) we sampled
\(n = 1000\) observations.

\hypertarget{sec:exps:performance:res}{%
\subsubsection{Results}\label{sec:exps:performance:res}}

\begin{figure}
\includegraphics[width=1\linewidth]{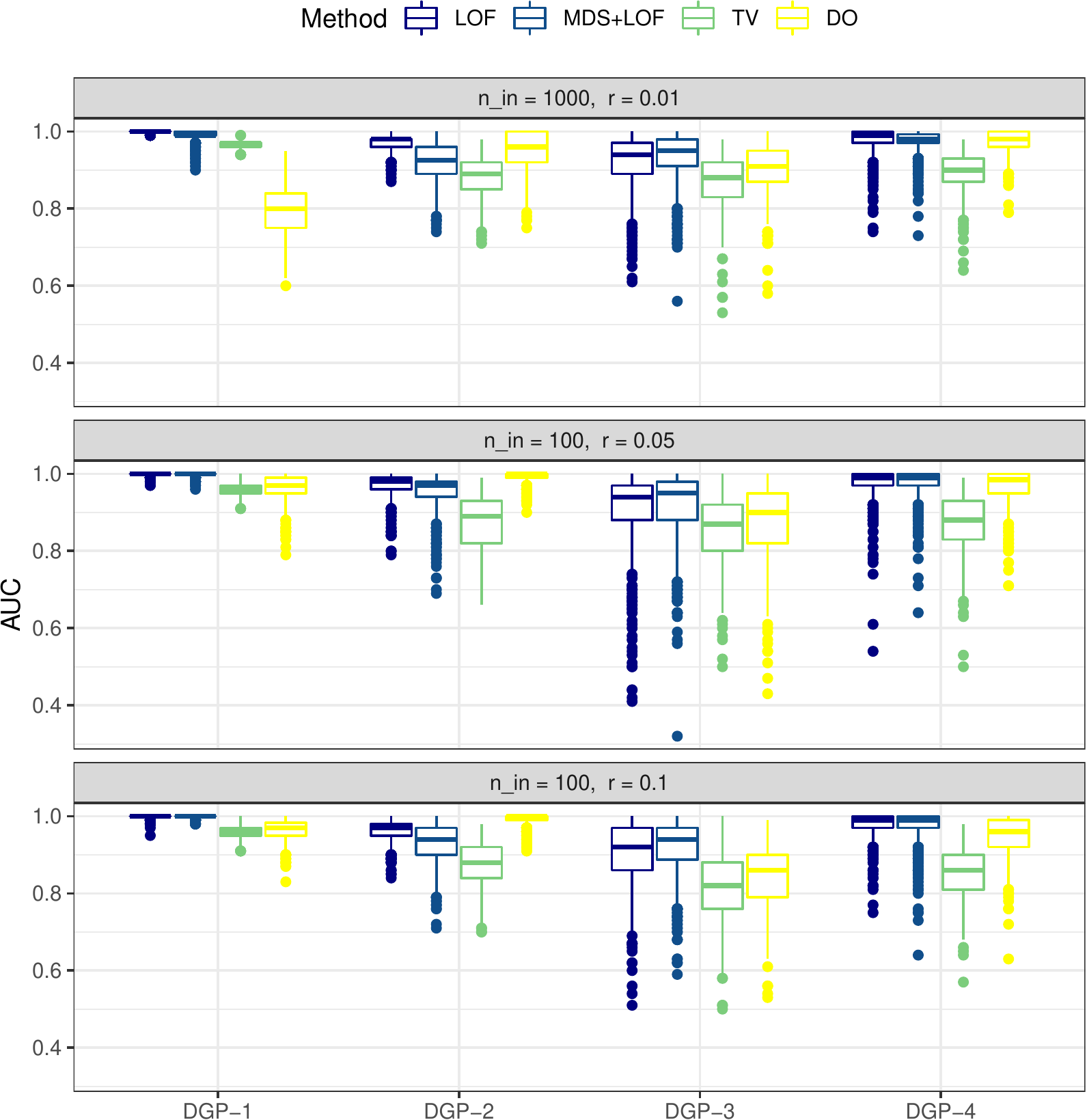} \caption{\label{fig:perf-complex}Distribution of AUC over the 500 replications for the different data generating processes (DGP), outlier detection methods, and outlier rations $r$.}\label{fig:exp-comp-perf}
\end{figure}

First of all, note that LOF directly applied to functional data as well
as LOF applied to 5D embeddings yield very similar results. This agrees
with our findings in the qualitative analyses. In the following, we
simply refer to the geometrical approach and do not distinguish between
LOF based on MDS embeddings and LOF applied directly to the functional
distance matrix. Figure \ref{fig:perf-complex} shows that the
geometrical approach is highly competitive with functional-data-specific
outlier detection methods. It yields better results than TV for all of
the four DGPs and performs at least on par with DO. In comparison to DO
it performs better on DGP 1 and DGP 3, on par on DGP 4, worse on DGP 2.
Note that DO struggles to detect simple shift outliers: of all methods
it performs worst on the first DGP. Similar conclusions can be reported
for other settings, where it performs even worse if there are only shift
outliers (c.f. Figures \ref{fig:perf-fdaoutlier},
\ref{fig:sms-compare}).\\
In summary, based on the conducted experiments the geometrical approach
leads to outlier scoring performances at least on par with specialized
functional outlier detection methods even if based on fairly basic
methods (MDS with \(L_2\) distances and LOF). Going further, our
approach can be adapted to specific settings simply by choosing metrics
other than \(L_2\). As the next section shows, this can improve the
outlier detection performance considerably.

\hypertarget{sec:exps:general-dists}{%
\subsection{General dissimilarity measures and manifold
methods}\label{sec:exps:general-dists}}

\begin{figure}[h]
\includegraphics[width=1\linewidth]{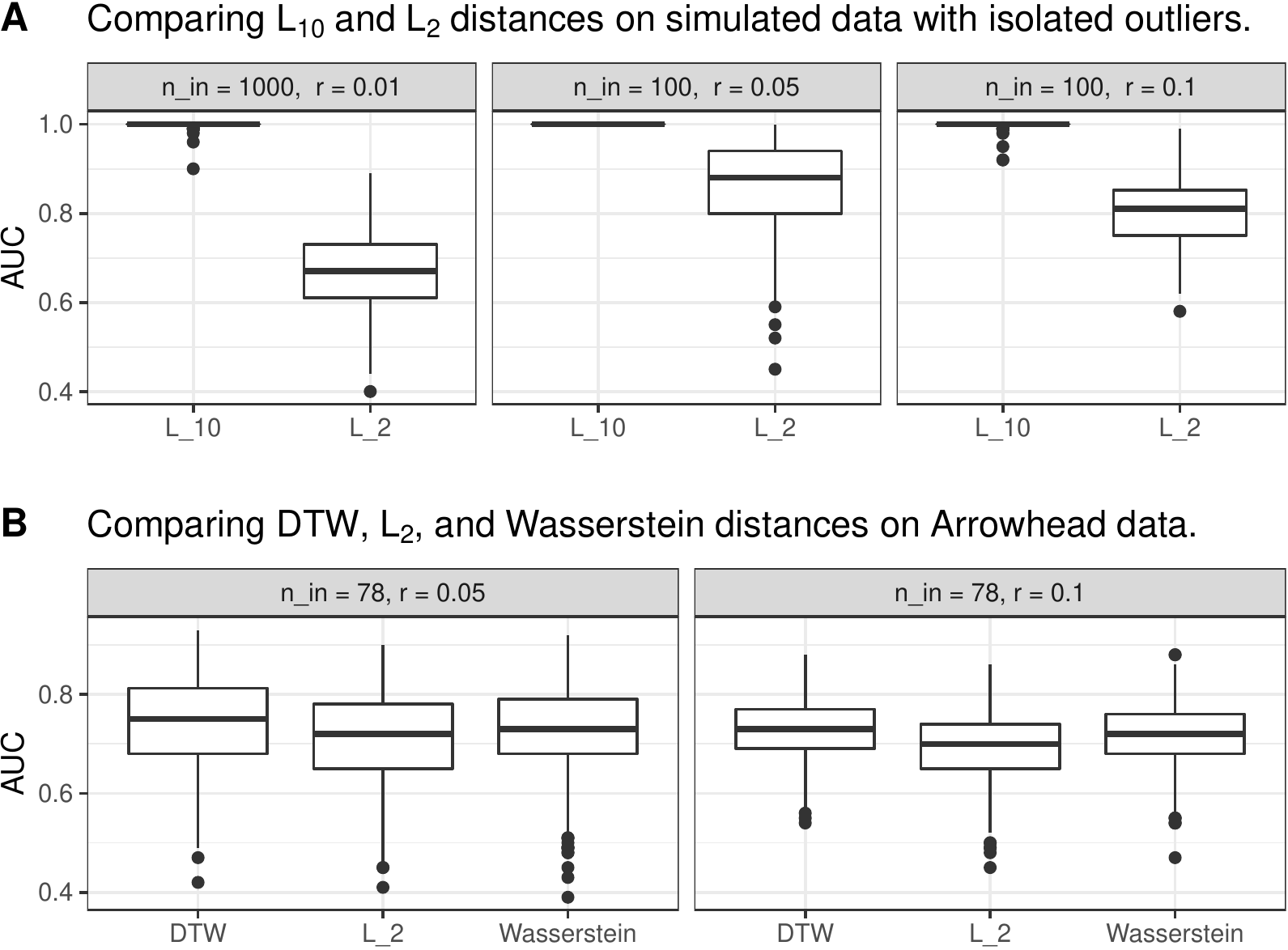} \caption{\label{fig:metric-vs}Comparing the effects of different distance measures. Depicted are the distributions of AUC over 500 replications for LOF based on MDS embeddings computed with the respective distance measures for different outlier ratios $r$. \textbf{A}: Comparing $L_{10}$ and $L_2$ metric on a data set with isolated outliers generated via simulation model 2 from package fdaoutlier. \textbf{B}: Comparing DTW, $L_2$, and unnormalized $L_1$-Wasserstein distance measures on real data set ArrowHead. Note: the DTW distance is not a  metric.}\label{fig:metric-plts}
\end{figure}

So far, we have computed MDS embeddings mostly based on \(L_2\)
distances. In the following we show that the approach is more general.
The geometric structure of a data set is captured in the matrix of
pairwise distances among observations. Different metrics emphasize
different aspects of differences in the data and can thus lead to
different geometries. MDS based on \(L_2\) distances yielded compelling
results in many of the examples considered above, but other distances
are likely to lead to better performance in certain settings. To
illustrate the effect, we consider two additional settings -- one
simulated and one on real data -- in the following. The results are
displayed in Figure \ref{fig:metric-vs}.\\
The simulated setting is based on isolated outliers, i.e.~observations
which deviate from functions in \(\Min\) only on small parts of their
domain. In such settings, higher order \(L_p\) metrics lead to better
results, since such metrics amplify the contribution of small segments
with large differences to the total distance. We use as an example data
generated from \(\verb|simulation_model2|\) from package
\(\verb|fdaoutlier|\). Figure \ref{fig:metric-vs} A shows the AUC values
of LOF scores on MDS embeddings based on \(L_2\) and \(L_{10}\)
distances. Again, 500 data sets were generated form the model over
different outlier ratios. In contrast to \(L_2\)-based MDS, using
\(L_{10}\) distances yields almost perfect detection. In embeddings
based on \(L_{10}\), isolated outliers are clearly separable in the
first two or three embedding dimensions. \\
As a second example, we consider the \textit{ArrowHead} data set
\citep{dau2019ucr, ye2009time}, which contains outlines of three
different types of neolithic arrowheads (see Appendix \ref{app:arrow}
for visualizations of the data set). Using the 78 structurally similar
observations from class ``Avonlea'' as our data on \(\Min\) and sampling
outliers from the 126 structurally similar observations from the other
two classes, we can compute AUC values based on the given class labels.
We generate 500 data sets for each outlier ratio
\(r \in \{0.05, 0.1\}\). Since there are only 78 observations in class
``Avonlea'', we do not use \(r = 0.01\) for this example. Embeddings are
computed using three different dissimilarity measures: the standard
\(L_2\) metric, the unnormalized \(L_1\)-Wasserstein metric
\citep{gangbo2019unnormalized}, and the Dynamic Time Warping (DTW)
distance \citep{rakthanmanon2012searching}. Note that the DTW distance
does not define a proper metric \citep{lemire2009faster}.\\

Figure \ref{fig:metric-vs} B shows that small performance improvements
can be achieved in this case if one uses dissimilarity measures that are
more appropriate for the comparison of shapes, but not as much as in the
isolated outlier example. Note that even though DTW distance is not a
proper metric, it improves outlier scoring performance in this example.
This indicates that, from a practical perspective, general dissimilarity
measures can be sufficient for our approach to work. This opens up
further possibilities, as there are many general dissimilarity measures
for functional data, for example the semi-metrics introduced by Fuchs et
al. \citep{fuchs2015nearest}. Overall, these examples illustrate the
generality of the approach: using suitable dissimilarity measures can
make the respective structural differences more easily distinguishable.

More complex embedding methods, on the other hand, do not necessarily
lead to better or even comparable results as MDS. Figure
\ref{fig:perf-uimap} shows the distribution of AUC for embedding methods
ISOMAP and UMAP. Both methods require a parameter that controls the
neighborhood size used to construct a nearest neighbor graph from which
the manifold structure of the data is inferred. The larger this value,
the more of the global structure is retained. For both methods,
embeddings were computed for very small and very large neighborhood
sizes of 5 and 90.\\
The results show that neither method performs better than MDS, UMAP even
performs considerably worse. Note that ISOMAP is equivalent to MDS based
on the geodesic distances derived from the nearest neighbor graph and
the larger the neighborhood size the more similar to direct pairwise
distances these geodesic distances become. This is also reflected in the
results, as ISOMAP-90 performs better than ISOMAP-5 on average. For
DGP-2, ISOMAP-90 slightly outperforms MDS, indicating that more complex
manifold methods could improve results somewhat in specific settings.\\
In general, however, these findings confirm the theoretical
considerations sketched in section \ref{sec:prelim:methods}. Embedding
methods which preserve the geometry of the space \(\funspace\) of which
\(\Min\) and \(\Man\) are sub-manifolds, i.e.~the ambient space
geometry, are more suited for outlier detection than methods which focus
on approximating the intrinsic geometry of the manifold(s). Thus, more
sophisticated embedding methods which often focus on approximating the
intrinsic geometry should not be applied lightly and certainly require
careful parameter selection in order to be applicable for outlier
detection. Since hyperparameter tuning for unsupervised methods remains
an unsolved problem, this is unlikely to be achieved in real-world
applications. In particular, consider that both UMAP and t-SNE
\citep{van2008visualizing} have been found to be -- in general --
oblivious to local density, which means that clusters of different
density in the observation space tend to become clusters of more equal
density in the embedding space \citep{narayan2021assessing}. Although
there may exist a parameter setting where this effect is reduced (note
that there are now density-preserving versions of t-SNE and and UMAP
\citep{narayan2021assessing}), we are skeptical that outliers can be
faithfully represented in such an embedding given the difficulties of
hyperparameter tuning in unsupervised settings. Moreover, these methods
are not designed to preserve important aspects of the outlier structure.
For example, UMAP is subject to a local connectivity constraint which
ensures that every observation is at least connected to its nearest
neighbor (in more technical terms: that a vertex in the fuzzy graph
approximating the manifold is connected by at least one edge with an
edge weight equal to one \citep{mcinnes2020umap}), which makes it
unlikely that UMAP can be tuned so that it is able to sensibly embed
off-manifold outliers, which should, by definition, not be connected to
the common data manifold. The poor performance of UMAP embeddings in our
experiments confirms these concerns.

\hypertarget{sec:dis}{%
\section{Discussion}\label{sec:dis}}

Based on a geometrical perspective of functional outlier detection, we
define two general types of functional outliers: off- and on-manifold
outliers. Our investigation shows that this perspective clarifies the
theoretical concepts and improves practical results. From a theoretical
perspective it allows to formalize functional outlier scenarios in
precise and consistent terms, beyond differences in terms of either
shape, level or magnitude. This simplifies reasoning about specific
outlier settings and provides a fully general theoretical
conceptualization of the problem.\\
From an applied perspective, we formulate two important consequences.
First of all, as has been demonstrated with a comprehensive analysis of
a complex, real data set of ECG curves, the geometrical approach allows
for easily accessible and highly informative visualizations. These are
obtained by means of low dimensional embeddings reflecting the inherent
structure of a functional data set in much detail. Such visualizations
provide more accurate and complete pictures of the (outlier) structure
of functional data. In particular, off-manifold outliers reliably appear
as clearly separated (groups of) points in the low dimensional
embeddings.\\
Second, the proposed approach makes it possible to apply
highly-developed and performant standard outlier detection methods to
functional data, since the geometric structure of the data is captured
and reflected in their pairwise distance matrices. Outlier detection and
scoring methods which can be applied to distance matrices directly can
therefore be used for functional data as well. Furthermore, detection
methods requiring tabular inputs can also be applied simply by using the
embedding coordinates obtained with embedding methods as proxy data for
the original functions. Our experiments using LOF scores show that the
two approaches yield very similar results. This simultaneously
simplifies and improves functional outlier detection: It simplifies,
since functional data analysis becomes more accessible to a broader
audience with general outlier detection methods that are widely used in
other areas and that do not require an understanding of complex
methodological details of functional data methods. It improves the state
of the art since many functional outlier methods can only detect
specific kinds of functional outliers by design, or fail in more complex
realistic data that are widely dispersed or that contain multiple
non-outlying subgroups like the ECG data. Moreover, note that our
proposal is not limited to univariate functional data. Extending it to
multivariate functions is completely straightforward, as long as a
suitable dissimilarity measure is available to compute pairwise
distances.\\
In this paper, most embeddings were obtained using MDS based on \(L_2\)
distances. This implies a close similarity to functional bagplots and
highest density region (HDR) boxplots \citep{hyndman2010rainbow}, which
are based on the first two robust principal component scores. However,
this similarity only applies if our geometrical approach is implemented
with 2D MDS embeddings based on \(L_2\) distances. As outlined, our
proposal is neither limited to the \(L_2\) metric as a distance measure
nor to MDS as an embedding method or just two embedding dimensions.
Other metrics and (higher-dimensional) embedding methods can be used and
the conducted experiments indicate that alternative distance measure can
further improve the performance in specific settings, sometimes
considerably. In particular, even non-metric dissimilarity measures may
be applicable as our results based on DTW distances indicate. On the
other hand, the results also show that more sophisticated embedding
methods such as ISOMAP and UMAP cannot be used as straightforwardly as
MDS. Such methods, which do not take into account the ambient space
geometry by default, at least require very careful parameter selection.
In terms of practical applicability, the \(O(n^3)\) time complexity and
\(O(n^2)\) storage complexity of standard MDS may prove problematic for
large data, but generalizations such as Landmark MDS
\citep{de2002global}, Pivot MDS \citep{brandes2006eigensolver} or
multilevel MDS exploiting GPU performance \citep{ingram2008glimmer}
scale much better with the number of available observations.\\
Finally, we would argue that existing functional outlier detection
approaches mostly lack the principled geometrical underpinning and
conceptualization presented here. As outlined, we argue that such a
conceptualization is necessary to make functional outlier detection
tractable in full generality. Specifically, consider that existing
methods typically limit themselves to creating a 1D or 2D representation
of each curve (e.g., MBD-MEI, MO-VO, functional bagplots, HDR plots),
often based on preconceived notions of the characteristics of functional
outliers. Our investigations and experiments suggest that this is often
not sufficient for real-world functional outlier detection: First, there
is no reason to limit representations to two dimensions with modern
outlier detection methods, and the geometrical perspective often
strongly suggests otherwise in the case of complex functional data. Even
more importantly, it is much more flexible to learn maximally
informative low dimensional representations directly from data instead
of starting with rather a rigid notion of which characteristics to look
at and to ignore the rest. The latter is likely to lead to results not
capturing the entire (outlier) structure of a given data set, which is
essential in real-world unsupervised settings and exploratory
analyses.\\
Based on theoretical considerations and the empirical results outlined
above, we conclude that the proposed approach is well suited for both
theoretical conceptualization and practical implementation of functional
outlier detection. In particular, the choice of embedding method should
consider whether it is able to preserve the extrinsic geometry of the
function space and simple MDS embeddings based on functional distances
provide a very strong baseline for that. On the basis of this work we
intend to further investigate the implications of the geometrical
perspective, such as the effects of other dissimilarity measures,
embedding and outlier detection methods, in future research.

\textit{Acknowledgements}\\
This work has been funded by the German Federal Ministry of Education
and Research (BMBF) under Grant No.~01IS18036A. The authors of this work
take full responsibility for its content.

\textit{Reproducibility}\\
All R code and data to fully reproduce the results are freely available
on GitHub: \url{https://github.com/HerrMo/fda-geo-out}.

\appendix

\newpage

\hypertarget{appendix}{%
\section{Appendix}\label{appendix}}

\hypertarget{sec:phase-var}{%
\subsection{Formalizing phase variation scenarios}\label{sec:phase-var}}

\begin{figure}[H]
\includegraphics[width=1\linewidth]{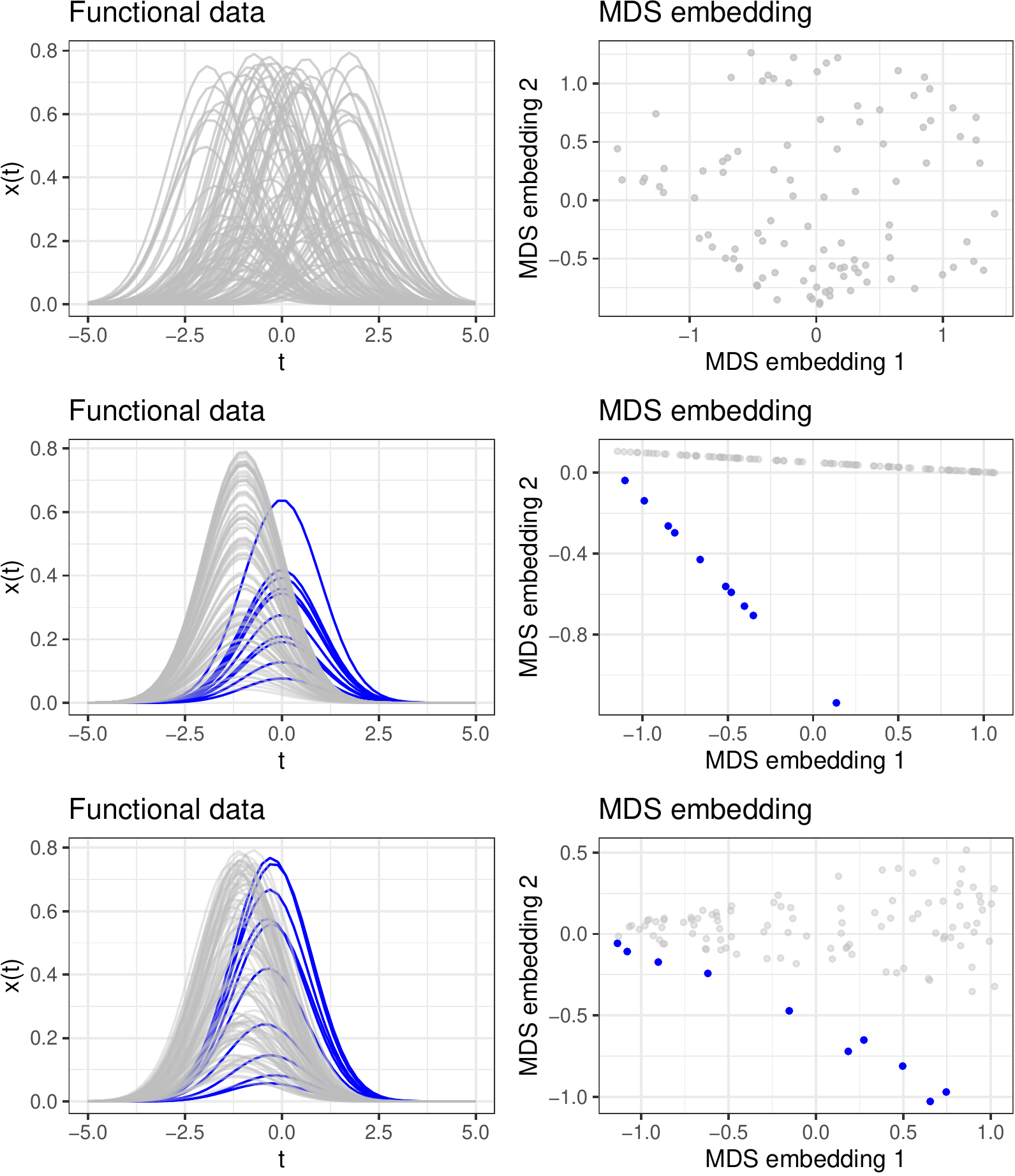} \caption{\label{fig:phase-example}Functional data with phase variation and different levels of structural difference. Top: Scenario with no off-manifold outliers. Mid: Scenario with clear off-manifold outliers. Bottom: Intermediate scenario.}\label{fig:unnamed-chunk-17}
\end{figure}

\textbf{Phase variation - Case I:} The manifold
\(\mathcal{M} = \{x(t): x(t) = \theta_1 \varphi(t - \theta_2), \mathbf{\theta} = (\theta_1, \theta_2)' \in \pspace\}\),
with \(\varphi(.)\) the standard Gaussian pdf and
\(\pspace = [0.1, 2] \times [-2, 2]\), defines a functional data setting
with independent amplitude and phase variation. Since there is a single
manifold only, there are no structural novelties. Figure
\ref{fig:phase-example} top depicts the functional observations on the
left and a 2D embedding obtained with MDS on the right. Note, all of the
curves are subject to amplitude and phase variation to varying extent,
however, there are no clearly ``outlying'' or ``outstanding''
observations in terms of either amplitude or phase. This is reflected in
the corresponding embedding, which does not show any clearly separated
observations in the embedding space, indicating that there are no
structurally different observations. The situation in the second case of
phase-varying data, however, is different.

\textbf{Phase variation - Case II:} The two manifolds
\(\Min = \{x(t): x(t) = \theta \varphi(t + 1), \theta \in \pspace\}\)
and \(\Man = \{x(t): x(t) = \theta \varphi(t), \theta \in \pspace\}\),
with \(\pspace = [0.1, 2]\), describe a similar scenario as before,
however, there are two structurally different manifolds induced by the
shift in the argument of \(\varphi\). In contrast to the first case,
there are on-manifold and off-manifold outliers. Figure
\ref{fig:phase-example} mid depicts the functional observations and the
corresponding embedding. Clearly, in this example few (blue) curves, the
ones from \(\Man\), show a horizontal shift compared to the normal data
and consequently those few curves appear horizontally ``outlying''.
Within the main data manifold, only on-manifold outliers in terms of
amplitude exist. These aspects are reflected in the corresponding
embedding: the low-dimensional representations of the blue curves are
clearly separated from those of the main data in grey.\\
Of course such clear settings -- in particular phase varying functional
data with fixed and distinct phase parameters -- will seldom be observed
in practice. A more realistic example is given by
\(\Min = \{x(t): x(t) = \theta_1 \varphi(t - \theta_2), (\theta_1, \theta_2)' \in \Thin\}\)
and
\(\Man = \{x(t): x(t) = \theta_1 \varphi(t - \theta_2), (\theta_1, \theta_2)' \in \Than\}\),
with \(\Thin = [0.1, 2] \times [-1.3, -0.7]\) and
\(\Than = [0.1, 2] \times [-0.5, 0.1]\). Here we have again two
structurally different manifolds. This is more realistic, since the
``phase parameters'' \(\theta_2\) are not fixed but are subject to
random fluctuations. In addition, the structural difference induced by
the phase parameters is much smaller. Considering Figure
\ref{fig:phase-example} bottom, again this is reflected in the
embedding: there are two separable structures, however the differences
are not as clear as in the second example above.\\
The three examples together show that the less similar the processes are
and/or the less variability there is within the phase parameters
defining the manifolds, the clearer structural differences induced by
horizontal variation become visible in the embeddings.

\hypertarget{sec:exps:sensitivity}{%
\subsection{Sensitivity analysis}\label{sec:exps:sensitivity}}

\begin{figure}[h]
\includegraphics[width=1\linewidth]{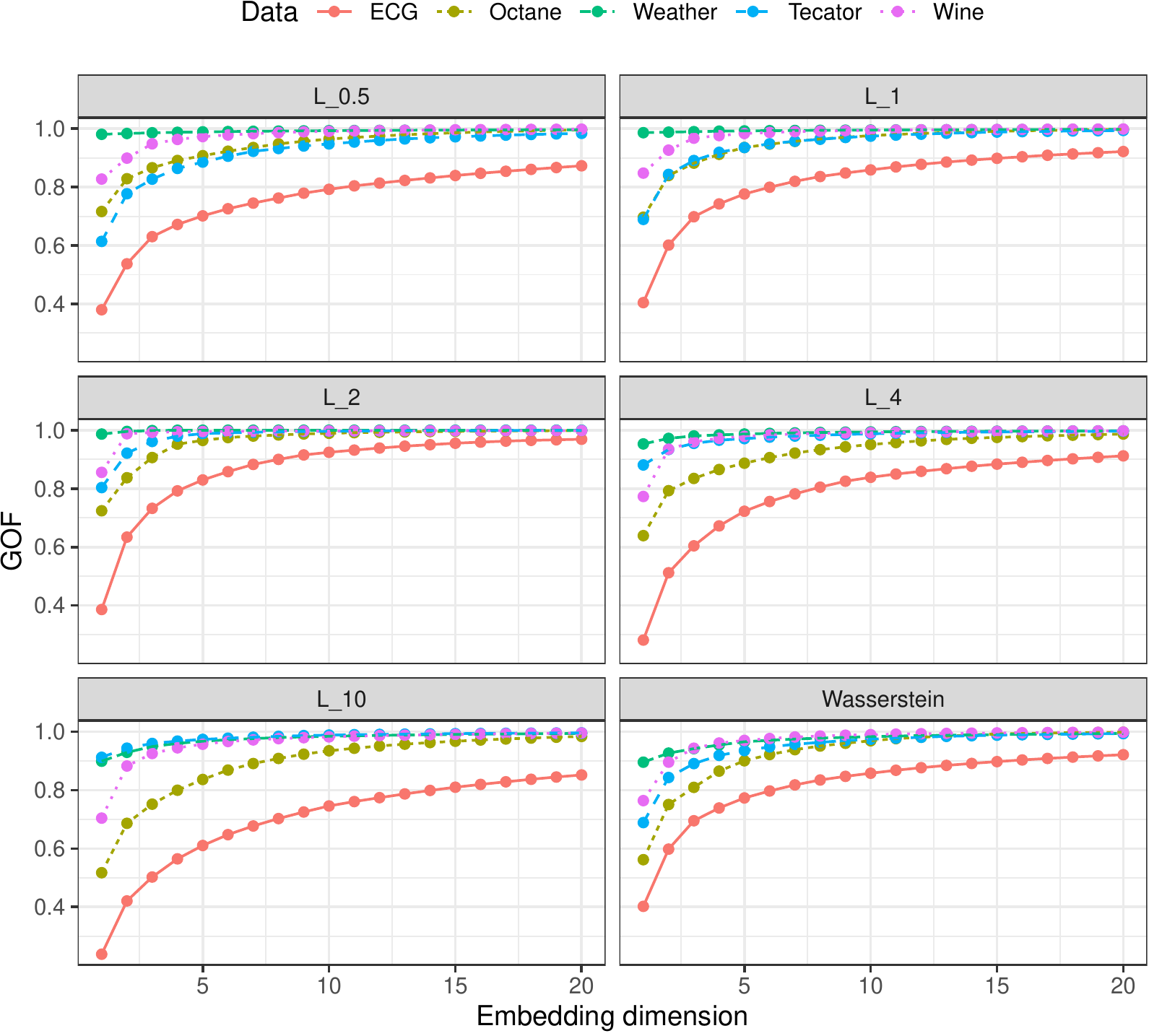} \caption{\label{fig:expl-variation} Goodness of fit (GOF) of different embedding dimensions for the five considered real data sets and $L_{0.5}, L_1, L_2, L_4, L_{10},$ and unnormalized $L_1$-Wasserstein metrics.}\label{fig:unnamed-chunk-18}
\end{figure}

The differences in complexity among the ECG and the other four real data
sets become apparent in Figure \ref{fig:expl-variation} as well, which
shows how the goodness of fit (GOF) of the embeddings is affected by
their dimensionality. For the \(L_2\) metric, a goodness of fit over 0.9
is achieved with two to three embedding dimensions for the less complex
data sets. Moreover, all of them reach a saturation point at five
dimensions. This is in contrast to the ECG data, where the first five
embedding dimensions lead to a goodness of fit of 0.8. Moreover, the
ranking induced by LOF scores is very robust to the number of embedding
dimensions. As Table \ref{tab:cor} shows, the rank correlations between
LOF scores based on five and LOF scores based on 20 embedding dimensions
are very high for all data sets.

\begin{table}[h]
\caption{\label{tab:cor}Spearman correlation between LOF scores based on embeddings of different dimensionality for the 5 considered real data sets and metrics $L_{0.5}$, $L_1$, $L_2$, $L_4$, $L_{10}$, and unnormalized $L_1$-Wasserstein. MDS embeddings with 5 dimensions are compared to embeddings with 2 (2vs5) and 20 (5vs20) dimensions.}
\centering
\begin{tabular}[t]{lrccccccccccc}
\toprule
\multicolumn{1}{c}{} & \multicolumn{2}{c}{$L_{0.5}$} & \multicolumn{2}{c}{$L_1$} & \multicolumn{2}{c}{$L_2$} & \multicolumn{2}{c}{$L_4$} & \multicolumn{2}{c}{$L_{10}$} & \multicolumn{2}{c}{Wasserstein} \\
\cmidrule(l{3pt}r{3pt}){2-3} \cmidrule(l{3pt}r{3pt}){4-5} \cmidrule(l{3pt}r{3pt}){6-7} \cmidrule(l{3pt}r{3pt}){8-9} \cmidrule(l{3pt}r{3pt}){10-11} \cmidrule(l{3pt}r{3pt}){12-13}
  & 2vs5 & 5vs20 & 2vs5 & 5vs20 & 2vs5 & 5vs20 & 2vs5 & 5vs20 & 2vs5 & 5vs20 & 2vs5 & 5vs20\\
\midrule
ECG & 0.96 & 0.97 & 0.98 & 0.97 & 0.97 & 0.99 & 0.94 & 0.98 & 0.85 & 0.94 & 0.98 & 0.97\\
Octane & 0.94 & 0.99 & 0.96 & 0.98 & 0.97 & 0.99 & 0.98 & 0.99 & 0.94 & 0.97 & 0.95 & 0.96\\
Weather & 1.00 & 1.00 & 1.00 & 1.00 & 1.00 & 1.00 & 1.00 & 1.00 & 1.00 & 1.00 & 1.00 & 1.00\\
Tecator & 0.97 & 0.99 & 0.96 & 0.99 & 0.99 & 1.00 & 0.99 & 1.00 & 0.99 & 1.00 & 0.96 & 0.99\\
Wine & 0.98 & 0.99 & 0.99 & 1.00 & 1.00 & 1.00 & 0.99 & 1.00 & 0.98 & 0.99 & 0.99 & 1.00\\
\bottomrule
\end{tabular}
\end{table}

\newpage

\hypertarget{sec:quant-fdaoutlier}{%
\subsection{\texorpdfstring{Quantitative results on \texttt{fdaoutlier}
package
DGPs}{Quantitative results on fdaoutlier package DGPs}}\label{sec:quant-fdaoutlier}}

\begin{figure}[h]
\includegraphics[width=1\linewidth]{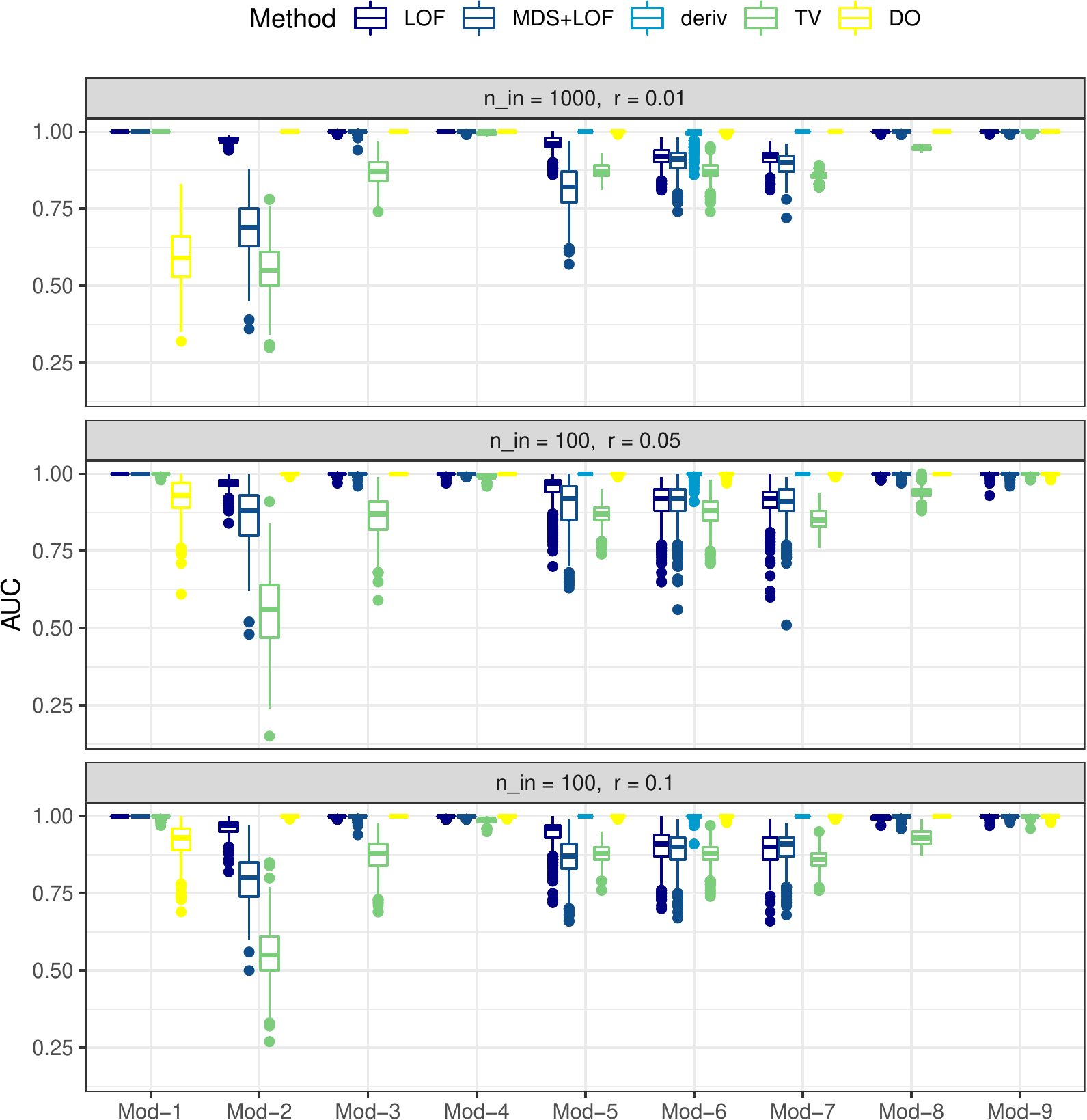} \caption{\label{fig:perf-fdaoutlier}Distribution of AUC over the 500 replications for the different outlier detection methods, simulation models (Mod) from package fdaoutier and outlier ratios $r$.}\label{fig:exp-fdaoutlier-perf}
\end{figure}

The simulation models presented by Ojo et al. \citep{ojo2021outlier}
cover different outlier scenarios: vertical shifts (model 1), isolated
outliers (model 2), partial magnitude outliers (model 3), phase outliers
(model 4), various kinds of shape outliers (models 5 - 8) and amplitude
outliers (model 9). A detailed description can be found in the
vignette\footnote{\url{https://cran.r-project.org/web/packages/fdaoutlier/vignettes/simulation_models.html}}
accompanying their R package. The same methods and performance
evaluation approach as in section \ref{sec:exps:performance} are used in
the following.\\
As Figure \ref{fig:perf-fdaoutlier} shows, (almost) perfect performance
is achieved by at least two methods for models 1, 3, 4, 8, and 9; DO
shows almost perfect performance for all models except model 1. For
models 2, 5, 6, and 7 the methods based on the geometric approaches do
not perform equally well (as does TV). However, as outlined in section
\ref{sec:exps:general-dists}, perfect performance can be achieved for
model 2 by using \(L_{10}\) distances instead of \(L_{2}\) distances.\\
Furthermore, for models 5, 6, and 7 it has to be taken into account that
the AUC values only reflect detection of ``true outliers'', which can
now -- given the geometric perspective -- be specified more precisely as
off-manifold outliers (observations from \(\Man\)). However, this does
not take into account possible on-manifold outliers. Due to their
distributional nature, by chance some on-manifold outliers (observations
on \(\Man\)) can be ``more outlying'' than some of the off-manifold
outliers and thus correctly obtain higher LOF scores. However, such
cases are not correctly reflected in the performance assessment
approach, as -- in contrast to off-manifold outliers -- such on-manifold
outliers are not labeled as ``true outliers''. The observed lower
performance in terms of AUC thus can simply mean that there are
on-manifold outliers obtaining relatively high LOF scores. In
particular, this also does not imply that off-manifold outliers fail to
be separated in a subspace of the embedding, as will be outlined
appendix \ref{sec:sms} in more detail, nor that perfect AUC performance
cannot be obtained via the geometric approaches for these settings. If
the geometric approach is applied to the derivatives instead (depicted
in Figure \ref{fig:perf-fdaoutlier} as ``deriv'') almost perfect
performances can be achieved. Obviously, functions of the same shape
(i.e.~all observations from \(\Min\)) are very similar on the level of
derivatives regardless of how strongly dispersed they are in terms of
vertical shift.

\hypertarget{sec:compare-outliergram}{%
\subsection{\texorpdfstring{Comparing embeddings,
\texttt{roahd::outliergram},
\texttt{fdaoutlier::msplot}}{Comparing embeddings, roahd::outliergram, fdaoutlier::msplot}}\label{sec:compare-outliergram}}

\begin{figure}[H]
\includegraphics[width=1\linewidth]{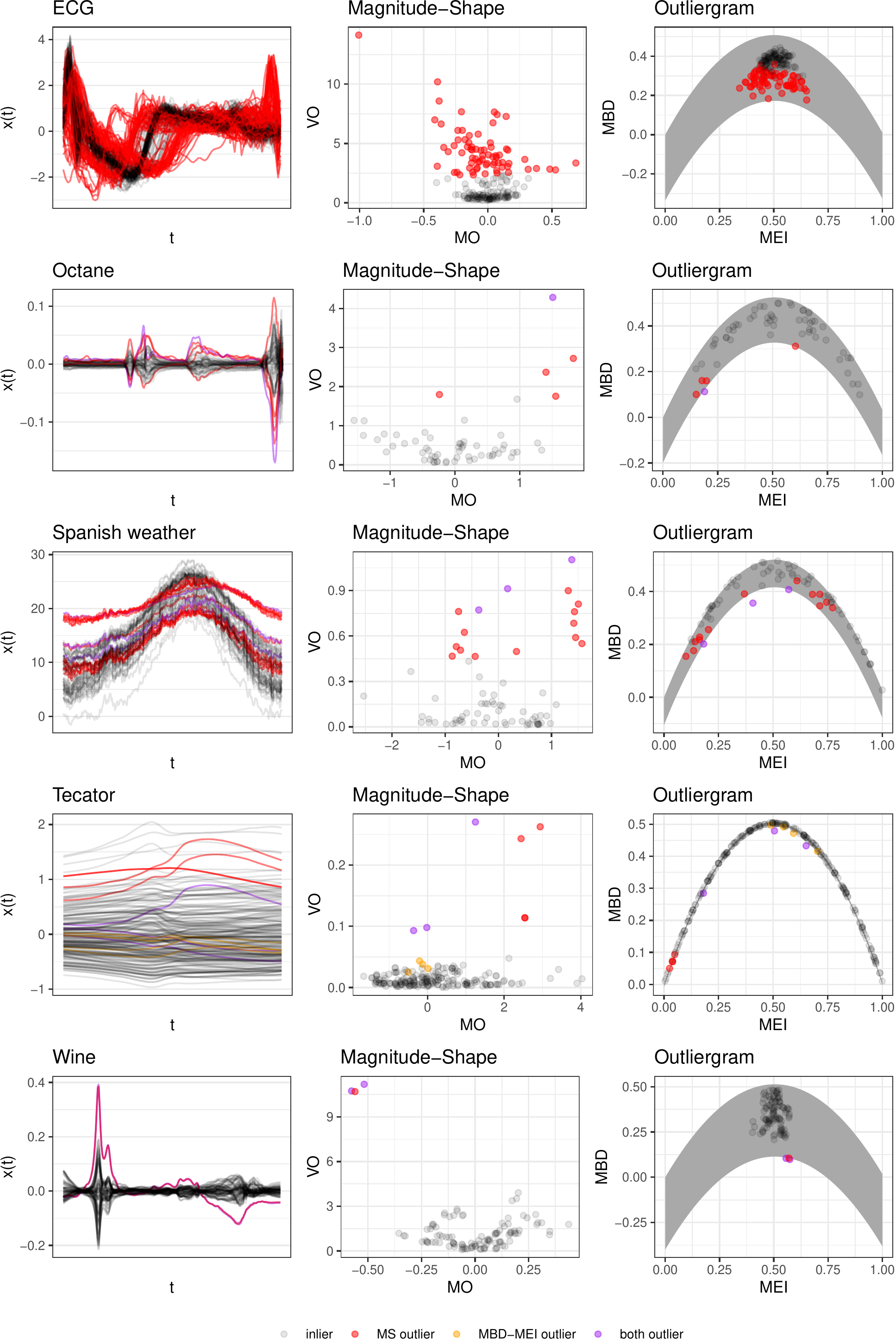} \caption{\label{fig:outliergram-real} Left column: data, middle column: magnitude-shape plots of mean directional outlyingness (MO) versus variability of directional outlyingness (VO), right column: Outliergram of modified epigraph index (MEI) versus modified band depth (MBD) with inlier region in grey. Curves and points are colored according to outlier status as diagnosed by \texttt{fdaoutlier::msplot} and/or \texttt{roahd::outliergram}.}\label{fig:fig-outliergram-real}
\end{figure}

Figure \ref{fig:outliergram-real} shows results for the MBD-MEI
``Outliergram'' by Aribas-Gil \& Romo
\citetext{\citealp{arribas2014shape}; \citealp[implementation:][]{roahd}}
for shape outlier detection and the magnitude-shape plot method of Dai
\& Genton \citep{dai2018multivariate} for the example data sets shown in
Figures \ref{fig:ecg} and \ref{fig:real-dat}.

Both of these visualization methods mostly fail to identify shift
outliers (by design, in the case of the outliergram). The outliergram
tends to mislabel very central observations as outliers in data sets
with little shape variability (e.g.~the ``shape outliers'' detected by
MBD-MEI in the central region of the Tecator data) and fails to detect
even egregious shape outliers in data sets with high variability
(e.g.~not a single MBD-MEI outlier for ECG) as well as shape outliers
that are also outlying in their level (e.g.~the 3 shape outliers
identified by \texttt{msplot} in the upper region of the Tecator data).
Note that some central functions of the Spanish weather data, which are
labeled as outliers by the magnitude-shape-plot (and partly by the
outliergram), are also reflected in the 2D embedding in Figure
\ref{fig:real-dat}. They are fairly numerous relative to the overall
sample size and are very similar to each other. As such, they form a
clearly defined separate cluster within the data, which can be seen in
the middle bottom part of the embedding.

\begin{figure}[h]
\includegraphics[width=1\linewidth]{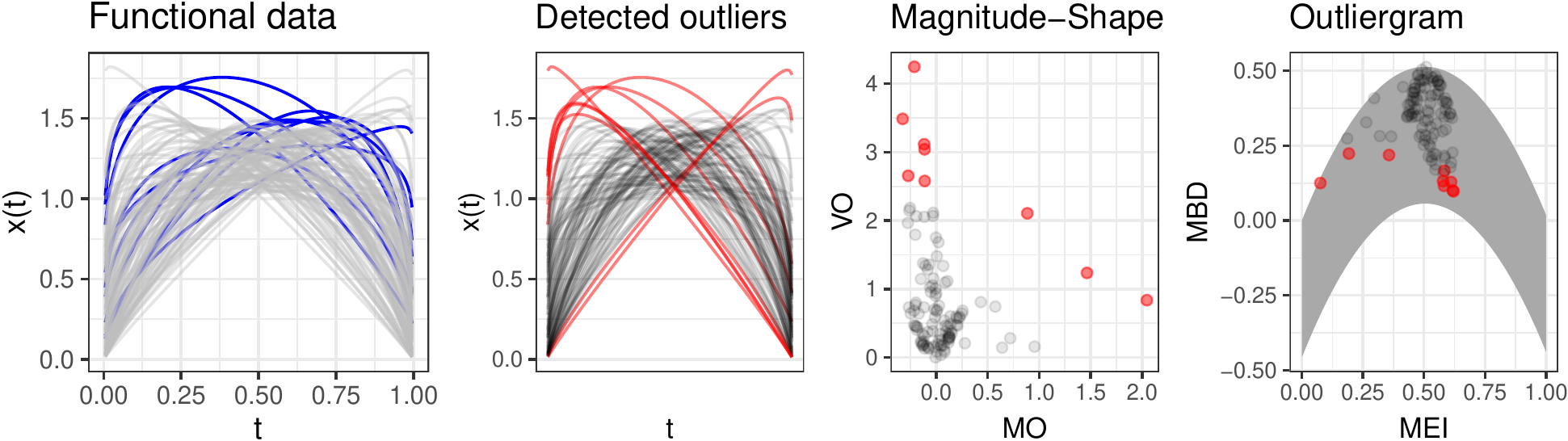} \caption{\label{fig:outliergram-ex5} From left: data with true outliers in blue, data with detected outliers in color, magnitude-shape plot of mean directional outlyingness (MO) versus variability of directional outlyingness (VO), outliergram of modified epigraph index (MEI) versus modified band depth (MBD) with inlier region in grey.}\label{fig:fig-outliergram-ex5}
\end{figure}

Figure \ref{fig:outliergram-ex5} shows the results for the synthetic
data example of Figure \ref{fig:shift-example} with 10 true outliers,
where the MS plot yields 6 false positives and only 3 true positives,
while the Outliergram fails to detect even a single outlier.

\newpage

\newpage

\hypertarget{sec:sms}{%
\subsection{In-depth analysis of simulation model 7}\label{sec:sms}}

\begin{figure}[h]
\includegraphics[width=1\linewidth]{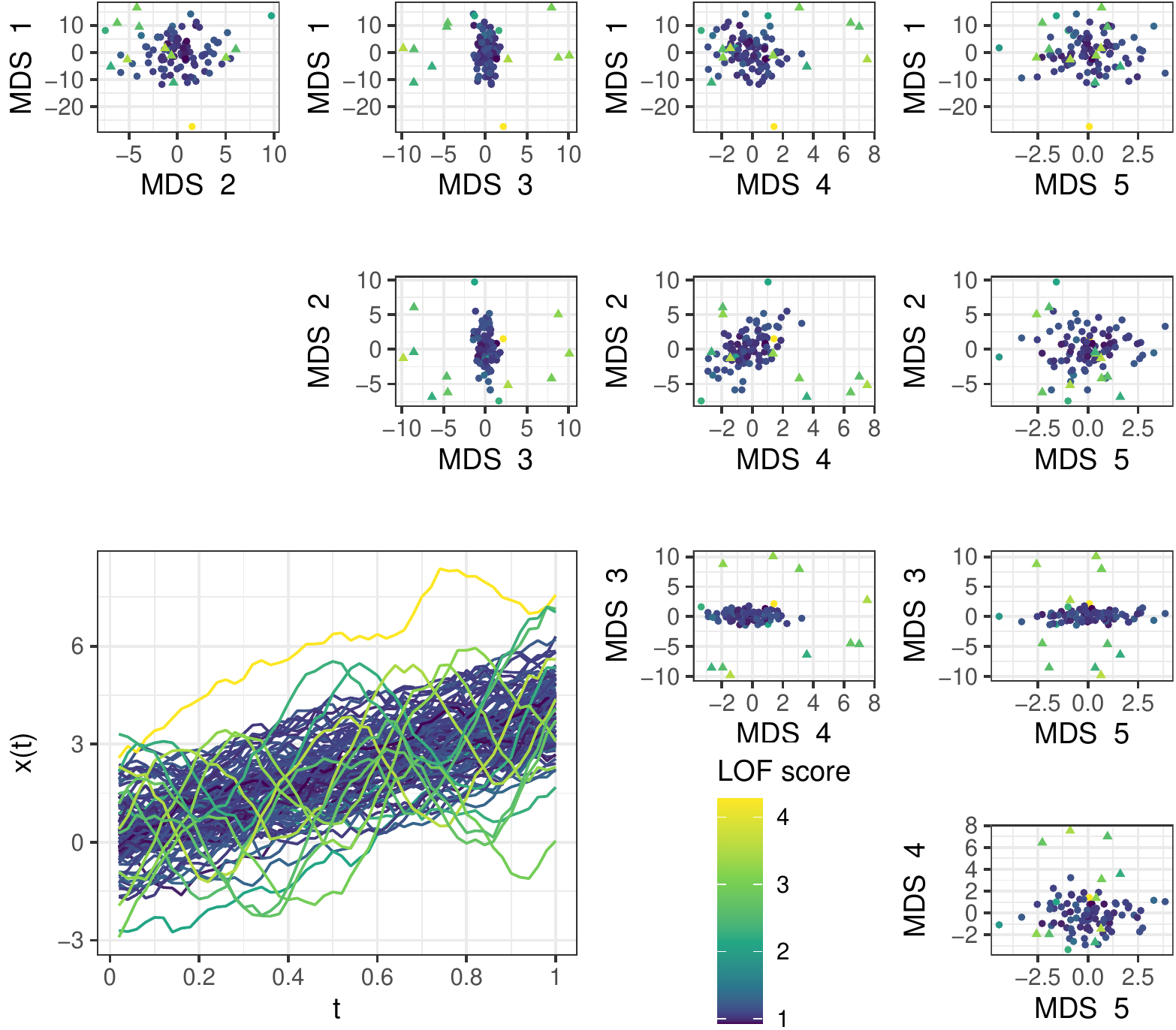} \caption{\label{fig:mod7}Model 7 data: Scatterplotmatrix of all 5 MDS embedding dimensions and curves, lighter colors for higher LOF score of 5D embeddings. True outliers depicted as triangles. Note that the true outliers are clearly separated from the rest of the data in embedding subspace 3vs4.}\label{fig:sms-scatter}
\end{figure}

The analysis of the ECG data in section \ref{sec:exps:qual-analysis} has
shown that embeddings can reveal much more (outlier) structure than can
be represented by scores and labels. To illustrate the effects described
in appendix \ref{sec:quant-fdaoutlier}, we conduct a similar qualitative
analysis for an example data set with observations sampled from
simulation model 7, see Figure \ref{fig:mod7}. The data set consists of
100 observations with 10 off-manifold or -- in more informal terms:
``true'' -- outliers. The functions are evaluated on 50 grid points. The
analysis shows that a quantitative performance assessment alone may
yield misleading results and again emphasizes the practical value of the
geometric perspective and low dimensional embeddings.

\begin{figure}
\includegraphics[width=1\linewidth]{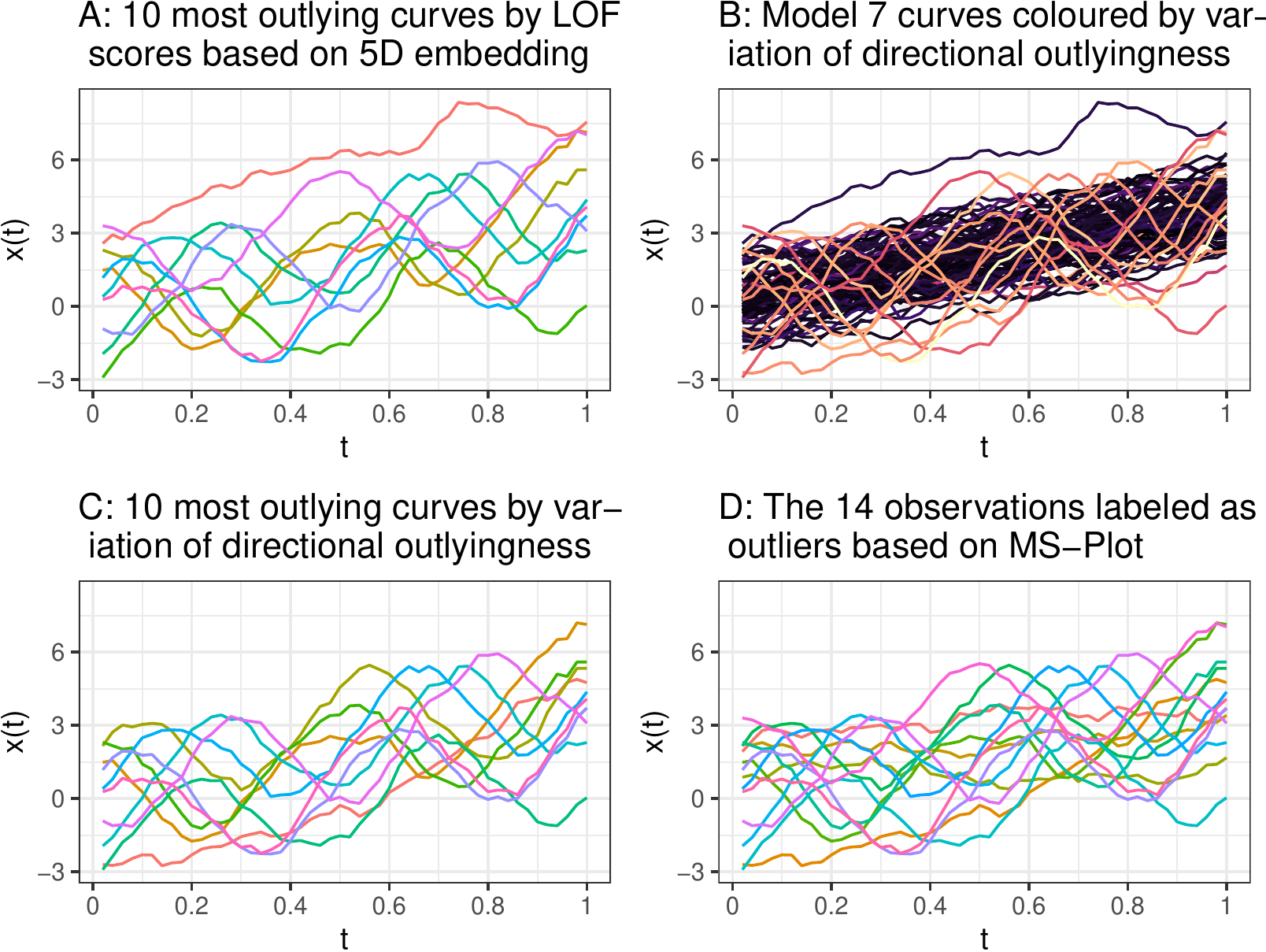} \caption{\label{fig:sms-compare}Model 7 data: LOF on MDS embeddings in contrast to directional outlyingness.}\label{fig:sms-funs}
\end{figure}

First of all, note that the AUC computed for this specific data set is
0.9, thus close to the median AUC value for LOF applied to MDS
embeddings of model 7 data, as depicted in Figure
\ref{fig:perf-fdaoutlier}. Nevertheless, the ``true outliers'' are
clearly separable in a 5D MDS embedding. As Figure \ref{fig:mod7} shows,
they are clearly separable in the subspace spanned by the third and
fourth embedding dimension. Note, moreover, that there is an outlying
observation with an extreme shift, which also obtains a high LOF score.
This observation is not labeled as a ``true outlier'' as it stems from
\(\Min\). This example shows that evaluation approaches for outlier
detection methods which are based on ``true outliers'' may not always
reflect the outlier structure adequately and may result in misleading
conclusions. However, those approaches are frequently used to compare
and assess different outlier detection methods. Again, this illustrates
the additional value low dimensional embeddings have for outlier
detection as such aspects become accessible.\\
Finally, note that DO/MS-plots are not sensitive to vertical shift
outliers as the extreme shift outlier is neither scored high based on DO
nor labeled as an outlier based on the MS-plot, see Figure
\ref{fig:sms-compare}.

\newpage

\hypertarget{app:dgp-ex}{%
\subsection{Examples for DGPs used for quantitative
evaluation}\label{app:dgp-ex}}

\begin{figure}[h]
\centering
\includegraphics{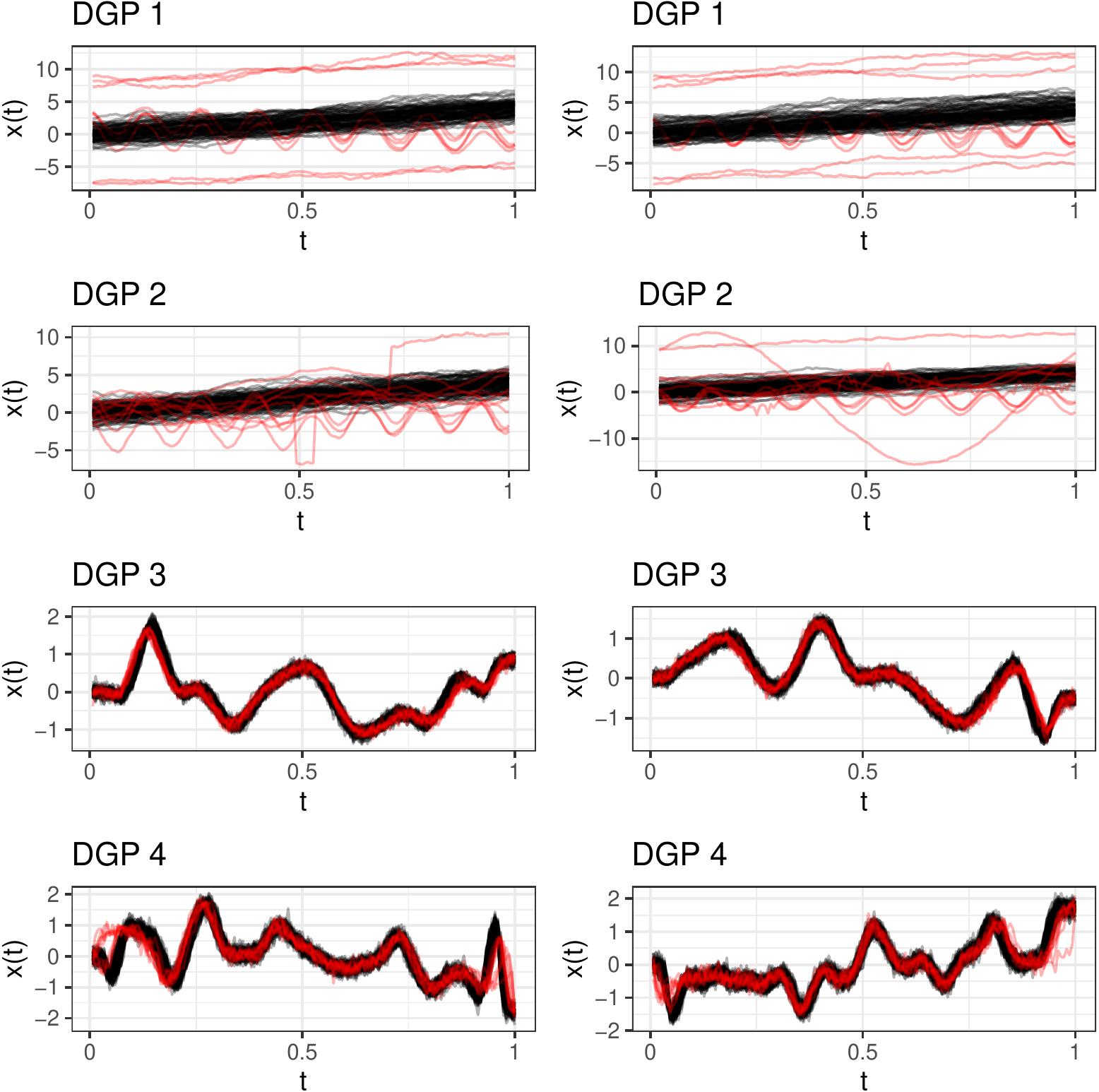}
\caption{\label{fig:dgp-examples} Example data sets for the DGPs used in
the simulation study (2 each). Inliers in black, outliers in red.
Outlier ratio 0.1, \(n = 100\).}
\end{figure}

Depicted in Figure \ref{fig:dgp-examples} are two example data sets for
each of the data generating processes (DGP) used in section
\ref{sec:exps:performance} for the comparison of the different outlier
detection methods.

\newpage

\hypertarget{app:arrow}{%
\subsection{Arrowhead data}\label{app:arrow}}

\begin{figure}[h]
\centering
\includegraphics{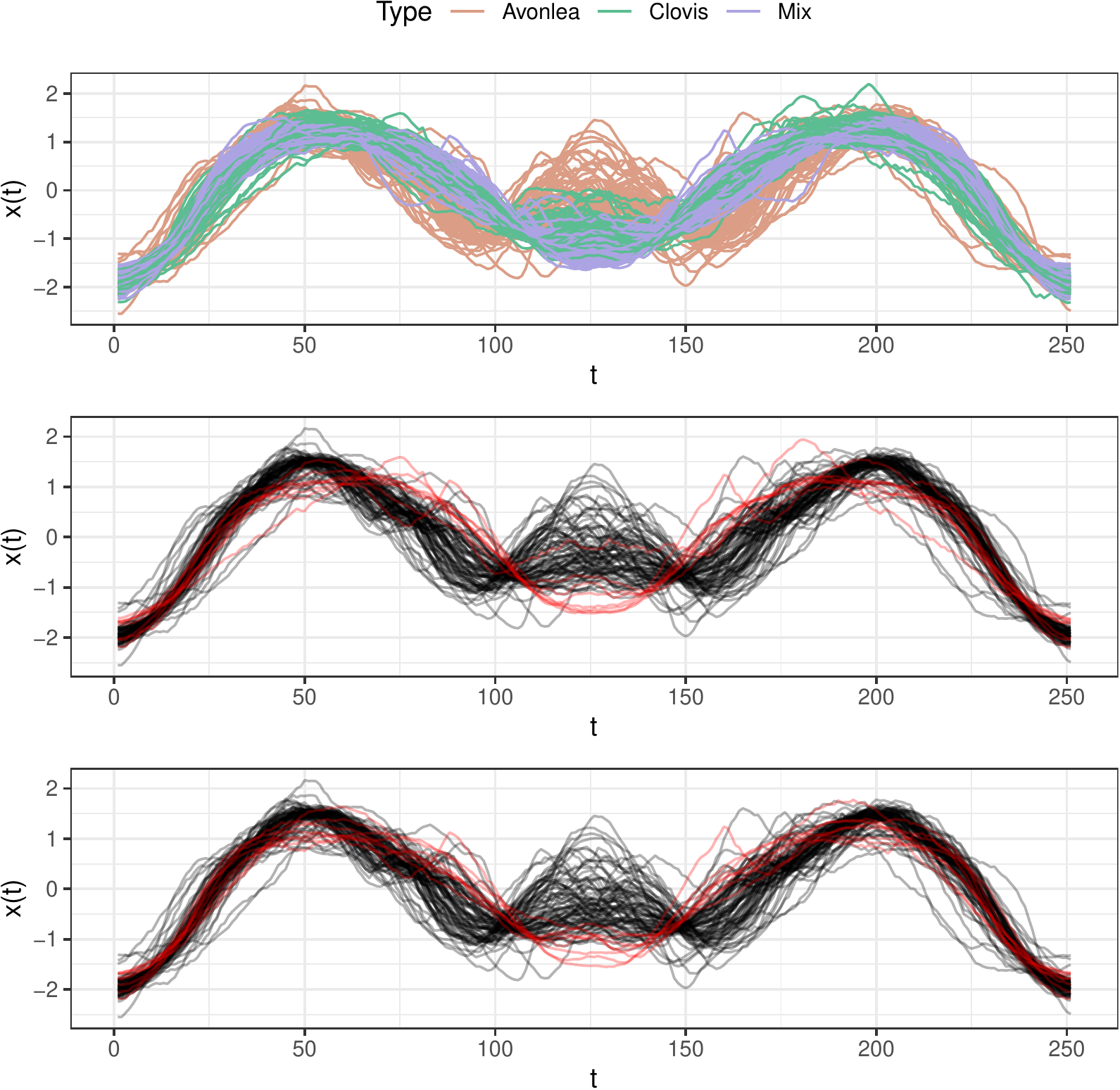}
\caption{\label{fig:arrowhead}ArrowHead data. Top: The complete data
set. Mid and bottom: Two example outlier data sets. Inliers from class
`Avonlea' in black, outliers sampled from classes `Clovis' and `Mix' in
red. Outlier ratio 0.1.}
\end{figure}

Depicted in Figure \ref{fig:arrowhead} are the ArrowHead data used in
section \ref{sec:exps:general-dists}.

\renewcommand\bibname{\leftline{References}}
\setlength{\bibsep}{0.0pt}

  \bibliography{references.bib}

\end{document}